%% file: preprint.tex
\documentclass{article}

\PassOptionsToPackage{numbers, compress}{natbib}

\usepackage[preprint]{neurips_2024}

\usepackage[utf8]{inputenc} 
\usepackage[T1]{fontenc}    
\usepackage{hyperref}       
\usepackage{url}            
\usepackage{booktabs}       
\usepackage{amsfonts}       
\usepackage{nicefrac}       
\usepackage{microtype}      
\usepackage{xcolor}         

\input{sections/shared_imports.tex}

\title{Intuitions of Compromise:\\ Utilitarianism vs. Contractualism}

\author{%
  Jared Moore \\
  Stanford University\\
  \texttt{jlcmoore@stanford.edu} \\
  \And
  Yejin Choi\\
  University of Washington\\
  \texttt{yejin@cs.washington.edu} \\  
  \\
  \And
  Sydney Levine \\
  Allen Institute for AI\\
  \texttt{sydneyl@allenai.org} \\
}

\begin{document}

\maketitle

\makeabstract

\maketitle

\newtoggle{pnas}
\togglefalse{pnas}

\input{sections/1_introduction}

\input{sections/2_related_work}

\input{sections/3_method+results}

\input{sections/5_discussion}
\input{sections/6_future_work}

\input{sections/7_conclusion}

{
\nocite{*}

\small
\bibliographystyle{plainnat}
\bibliography{value_aggregation, manual}

}

\appendix

\section{Supporting Information}

\input{sections/8_appendix}

\clearpage

\counterwithin{figure}{section}

\section{Supporting Information Results}

\input{sections/appendix_figures}

\end{document}

%% file: sections/shared_imports.tex
\usepackage[utf8]{inputenc} 
\usepackage[T1]{fontenc}    
\usepackage{hyperref}       
\usepackage{url}            
\usepackage{booktabs}       
\usepackage{amsfonts}       
\usepackage{nicefrac}       
\usepackage{microtype}      
\usepackage{xcolor}         
\usepackage{amsmath}
\usepackage{graphicx}
\usepackage{multirow} 
\usepackage{chngcntr} 
\usepackage{soul} 
\usepackage{pdfpages} 

\usepackage [english]{babel}
\usepackage [autostyle, english = american]{csquotes}
\MakeOuterQuote{"}

\newtoggle{comments}
\toggletrue{comments}

\iftoggle{comments}{
\newcommand{\jared}[1]{\textcolor{blue}{[Jared: #1]}}
\newcommand{\sydney}[1]{\textcolor{teal}{[Sydney: #1]}}
}{
\newcommand{\jared}[1]{}
\newcommand{\sydney}[1]{}
}

\DeclareMathOperator*{\argmax}{arg\,max} 

\usepackage{scalerel} 

\def\apple{\scalerel*{\includegraphics{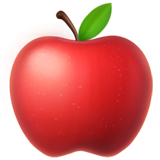}}{X}}
\def\cow{\scalerel*{\includegraphics{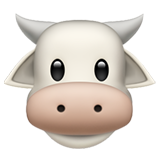}}{X}}
\def\bee{\scalerel*{\includegraphics{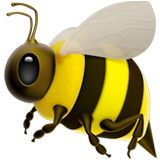}}{X}}

\usepackage{alltt} 
\usepackage{tcolorbox} 

\definecolor{areacolor}{rgb}{0.5529411764705883, 0.6274509803921569, 0.796078431372549}
\definecolor{areaaccent}{rgb}{0.4529411764705883, 0.5274509803921569, 0.696078431372549}
\definecolor{volumecolor}{rgb}{0.9058823529411765, 0.5411764705882353, 0.7647058823529411}
\definecolor{volumeaccent}{rgb}{0.8058823529411765, 0.4411764705882353, 0.6647058823529411}

\newtcolorbox{humanbox}[1]{title=#1}
\newtcolorbox{areabox}{colback=areacolor, colframe=areaaccent, title=Area Condition}
\newtcolorbox{volumebox}{colback=volumecolor, colframe=volumeaccent, title=Volume Condition}

\newcommand{\makeabstract}{
\begin{abstract}

What is the best compromise in a situation where different people value different things?  The most commonly accepted method for answering this question---in fields across the behavioral and social sciences, decision theory, philosophy, and artificial intelligence development---is simply to add up utilities associated with the different options and pick the solution with the largest sum.  This ``utilitarian'' approach seems like the obvious, theory-neutral way of approaching the problem.  But there is an important, though often-ignored, alternative: a ``contractualist'' approach, which advocates for an agreement-driven method of deciding. Remarkably, no research has presented empirical evidence directly comparing the intuitive plausibility of these two approaches. In this paper, we systematically explore the proposals suggested by each algorithm (the ``Utilitarian Sum'' and the contractualist ''Nash Product''), using a paradigm that applies those algorithms to aggregating preferences across groups in a social decision-making context. While the dominant approach to value aggregation up to now has been utilitarian, we find that people strongly prefer the aggregations recommended by the contractualist algorithm. Finally, we compare the judgments of large language models (LLMs) to that of our (human) participants, finding important misalignment between model and human preferences.

\end{abstract}
}

\newcommand{\scenarioCaption}{
\caption{\textbf{A}: An example scenario. We asked participants to choose between three proposals which would differentially effect three equally-sized groups. In this case, each proposal decreases the average cost of a medical visit. 
We either showed participants just the text on the left (\texttt{none} of the charts) or some combination of charts (\texttt{area}, \texttt{volume}, or \texttt{both}) to aid understanding of the scenarios. We kept the color of the proposals the same in both charts. \\[.25\baselineskip]
\textbf{B}: A \textbf{3-d, volume chart} of the scenario depicted in panel A. Each of the lines labelled "apple", "bee", and "cow" is an axis for each group. The colored boxes "one", "two", and "three" represent the different proposals. Each proposal spans a length on each axis proportional to the outcome for that group. (E.g. The blue box, "one" spans 1 on the "apple" axis, 51 on the "bee" axis, and 51 on the "cow" axis.) These 3-d charts could be dragged around with a cursor to see the boxes from different sides. 
We tested for this behavior and extensively familiarized participants with these 3-d charts in a qualification task (reproduced in SI \S~"Qualification Task"). \\[.25\baselineskip]
\textbf{C}: A \textbf{stacked, area chart} of the  scenario depicted in panel A. Each group appears on the x-axis. The colored bars show the outcome for each proposal for each group. }
\label{fig:scenario-charts}
}

\newcommand{\figureScenarioPNAS}{
\begin{SCfigure*}[\sidecaptionrelwidth][t!]
\centering
\includegraphics[width=.67\textwidth]{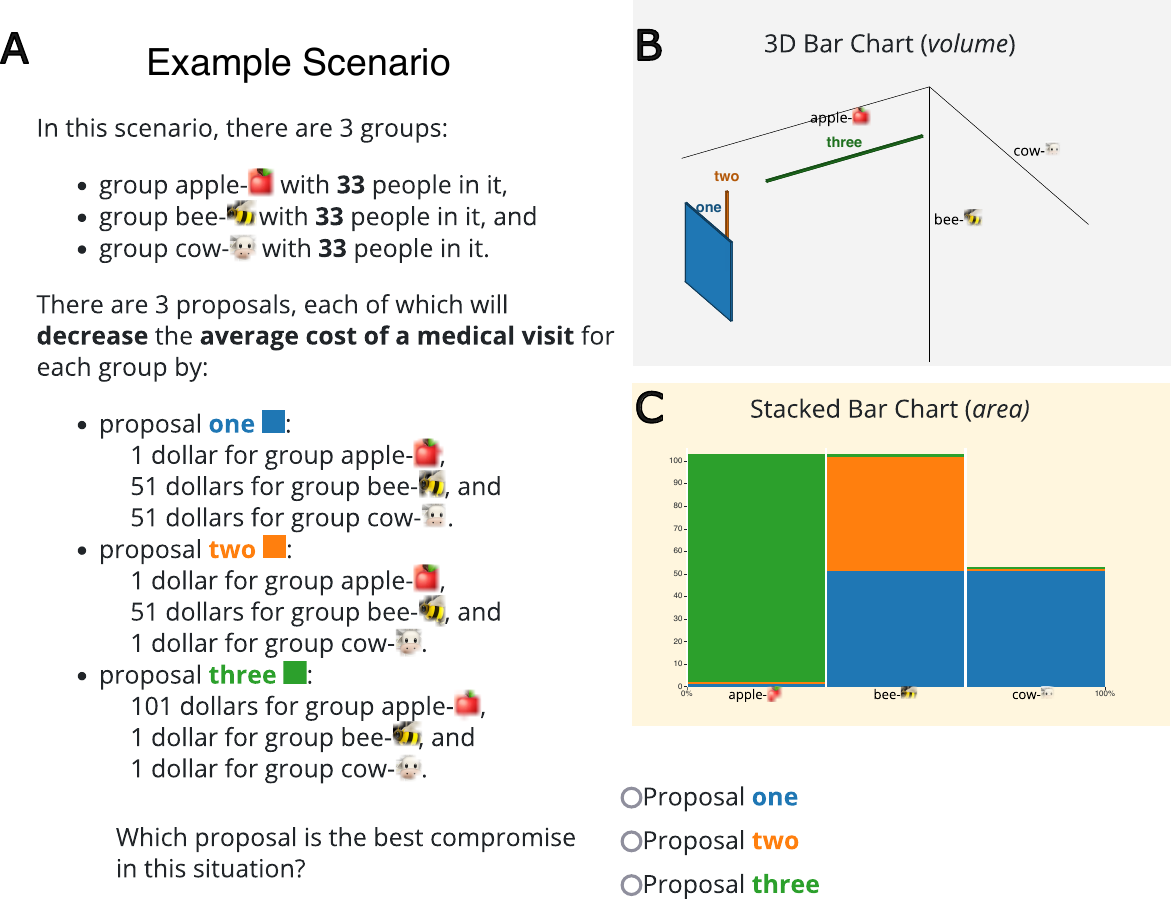}
\scenarioCaption
\end{SCfigure*}

}

\newcommand{\figureScenario}{
\begin{figure*}[t!]
\centering
\includegraphics[width=\textwidth]{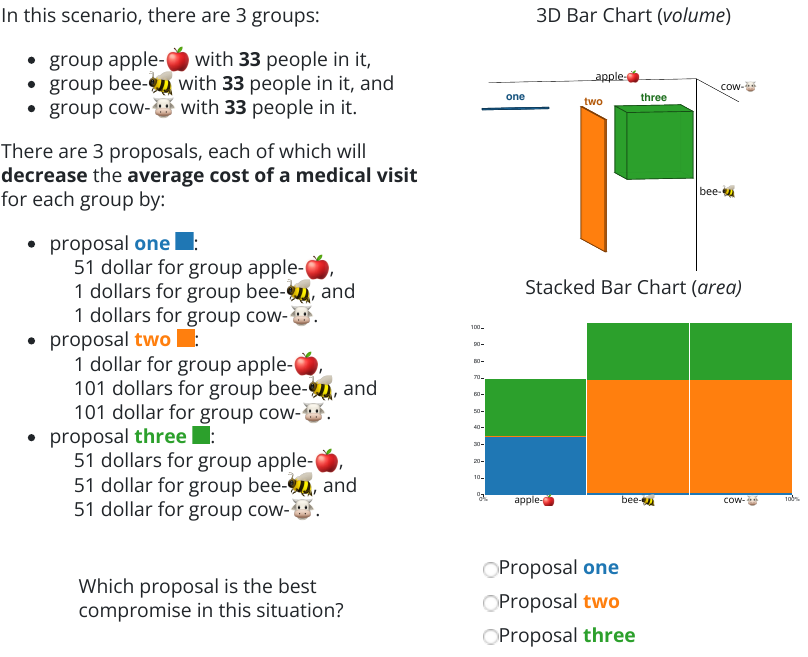}
\scenarioCaption
\end{figure*}
}

\newcommand{\figureMturkGptTest}{
\begin{figure}[t!]
\centering
\includegraphics[width=1\columnwidth]{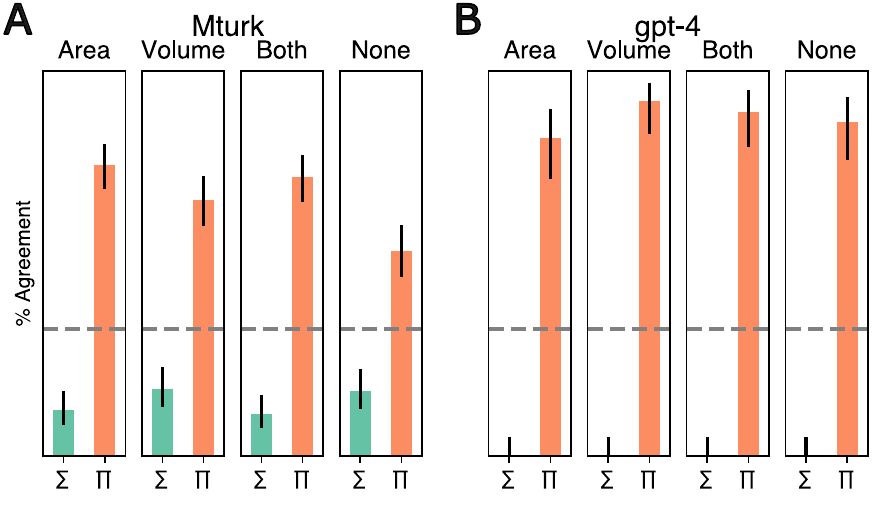}
\caption{The percent of \textbf{A} human (Mturk) participants and \textbf{B} \texttt{gpt-4} responses that endorse each value aggregation algorithm: the Utilitarian Sum (an additive model, shown in green with the $\Sigma$ symbol) and the Nash Product (a multiplicative model, shown in orange with the $\Pi$ symbol) on cases in which the two mechanisms disagree.  The panels represent the different decision-aids that participants received: \texttt{area}, \texttt{volume}, \texttt{both}, and \texttt{none}. (N=102 per condition.) The dashed line at 33\% indicates random guessing. (Participants always selected from three options.) Error bars show 95\% binomial confidence intervals. See SI Fig. A.4 for tests with additional LLMs.}
\label{fig:test-conditions}
\end{figure}
}

\newcommand{\tableMturk}{
\begin{figure}[t!]
{\centering
\setlength\tabcolsep{3pt}  
\begin{tabular}{r || r | c | c | c | c }
 &&  \multicolumn{4}{c}{Condition (\# of aligned responses / total \#)} \\ 
&  & \texttt{area} & \texttt{volume} & \texttt{both} & \texttt{none} \\ \hline \hline
Models & $\Pi$ &  165 / 216 & 145 / 216 & 158 / 216 & 116 / 216\\
\textbf{Disagree} &  & *** & *** &  ***& *** \\
& $\Sigma$ &  26 / 216 & 38 / 216 & 24 / 216 & 37 / 216 \\
&  &  *** &  *** &  *** & *** \\ \hline 
Models & $\Pi$ \& $\Sigma$ &  97 / 132 & 85 / 132 & 100 / 132 & 56 / 132  \\
\textbf{Agree} &  & *** & *** & *** & *
\end{tabular}

}

\caption{
The count of aligned \textbf{human} participants and each of the Nash Product ($\Pi$) and the Utilitarian Sum ($\Sigma$) when those mechanisms disagreed with each other and when they agreed. 
(See Fig. \ref{fig:test-conditions} and \ref{fig:mturk-control}.)
Columns show the visual aids participants received: the \texttt{area} chart, \texttt{volume} chart, \texttt{both}, or \texttt{none}. (N=102 per cell.)
The disagreement cases contained 18 unique scenarios presented with 4 different contexts each answered by 3 unique participants for  216 responses total ($18 \times 4 \times 3$). Similarly, the agreement cases had  132 responses ($11 \times 4 \times 3$).
In each case, we run a binomial test with a null hypothesis of random guessing (1/3). $\text{***}: p < .001$; $\text{*}: p < .05$}
\label{tab:results-mturk}
\end{figure}
}

\newcommand{\tablePrevalence}{
\begin{figure}[t]
\centering
\begin{tabular}{r| c | c}
& \% Disagree & \# Disagree / n \\ \hline 
Our generations -- $\{1, 51, 101\}$ & .82 & 162 / 19657 \\
Our generations -- $[1 \ldots 101]\spadesuit$ & 17 & 172144 / 999901 \\
Value Kaleidoscope \citep{sorensen_value_2023} & 15 &  1521 / 98694\\ 
NLPositionality \citep{santy_nlpositionality_2023} & 1.0 & 3 / 291 \\
Moral Machines \citep{awad_moral_2018} & .7 & 89 / 12600 
\end{tabular}
\caption{Structuring various data sets into the assumptions required to use aggregation mechanisms, we find that disagreements between the Utilitarian Sum and the Nash Product arise naturally. These figures should be interpreted as ballparks; given the numerical character of the Utilitarian Sum and the Nash Product, the number of disagreements varies dramatically with the shape of the numerical input. "Our generations -- $\{1, 51, 101\}$" are the Focused scenarios. "Our generations -- $[1 \ldots 101]$" are the Random scenarios. ($\spadesuit$ averages three samples.)  We describe both at the end of \S~"Scenario Generation."  See SI \S~"Prevalence" for more explanation and SI Fig. A.12 for an example.}
\label{tab:prevalence}
\end{figure}
}

\newcommand{\figureMturkLlmControl}{
\begin{figure}
\includegraphics[width=1\columnwidth]{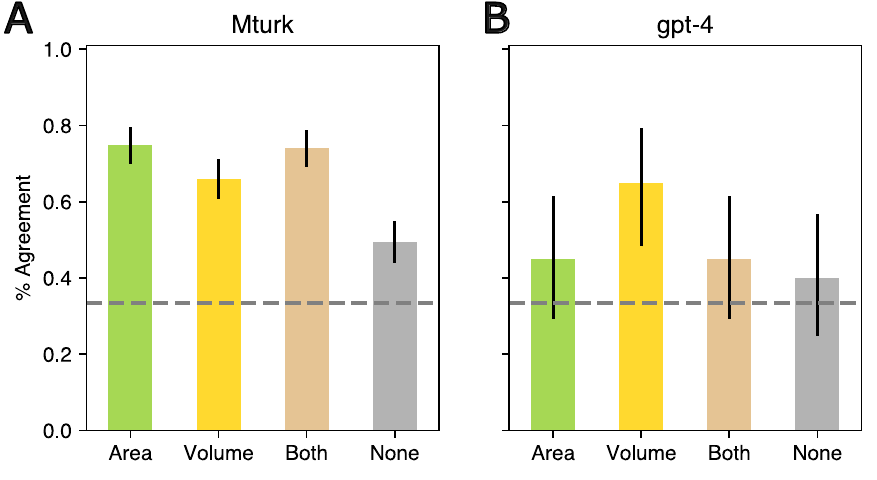}

\caption{The percent of \textbf{A} {human} participants (Mturk) and \textbf{B} {\texttt{gpt-4}} responses that were aligned with the Utilitarian Sum and the Nash Product on cases in which the two mechanisms agree. The panels represent the decision-aids participants received: \texttt{area}, \texttt{volume}, \texttt{both}, and \texttt{none}.  (N=102 per condition.)  The dashed line at 33\% indicates random guessing. \\[.25\baselineskip]
\textbf{A}: High agreement with the Utilitarian Sum and the Nash Product when both agree indicates that the two capture what participants intuit by a "best compromise."  \\[.25\baselineskip]
\textbf{B}: In comparison to the human results, the lower agreement of LLMs with the Utilitarian Sum and the Nash Product  when both agree indicates that computations besides those mechanisms drive the choice of a "best compromise."  This figure displays results for \texttt{gpt-4}; see SI Fig. A.5 for tests with additional LLMs.}
\label{fig:mturk-control}
\label{fig:llm-control}
\end{figure}
}

%% file: sections/1_introduction.tex
\iftoggle{pnas}{
\dropcap{I}magine
}{Imagine}
you lead a grant-making body answerable to groups of constituents with diverse needs and interests.  How should you allocate your limited funds?  Imagine you could give the money to a group of teachers who want to buy art supplies for their classrooms, a group of residents who want to organize a block party for the whole town, or a group of municipal workers looking for funds to help their colleagues unionize.  One thing you might decide to do is allocate all the money to the group that makes the most compelling argument.  In many cases, however, you would instead consider looking for some kind of compromise.  You could allocate funds in proportion to the number of constituents in each group and the magnitude of their need---this could result in maximizing overall welfare. Or you could call the constituents together to discuss the issue---this could result in determining a solution that all the invested parties would agree to.

These are two prominent ways of thinking about compromise when faced with the challenge of \emph{value aggregation}.  How should limited resources be distributed when different people value different things?  Two major schools of thought have competing proposals.  The ``utilitarian'' approach advocates for simply adding up utilities associated with everyone's welfare and picking the solution with the largest sum (Equation~\ref{equation-sum}).  In contrast, a ``contractualist'' approach advocates for an agreement-driven method of deciding.  There are a range of contractualist proposals, but here we focus on one commonly used formulation: the Nash Product (Equation~\ref{equation-product}).

Despite there being (at least) two theoretically-motivated approaches to the problem of value aggregation, in practice, research across fields from decision theory \citep{von_neumann_theory_1947, luce_games_1957}, to AI \citep{chen_just_2020, awad_moral_2018, gordon_jury_2022, sorensen_value_2023},
to philosophy \citep{lockhart_moral_2000, ross_rejecting_2006, sepielli_what_2014} have operated (often unreflectively) using the utilitarian approach. Moreover, to our knowledge, there has been little if any empirical investigation of which approach yields more intuitively plausible results.

We empirically survey participants' intuitions about the recommendations given by these contrasting approaches. Unlike most past work, we randomly generate and sample the proposals suggested by each mechanism instead of looking at isolated, illustrative cases. In addition, we design a series of visual aids to convey the proposals to participants.  This allows us to use quantitatively precise stimuli, while not overwhelming subjects with task-intensive, numerical comparisons.  Finally, we test the alignment of large language models (LLMs) to the judgments of our (human) participants to investigate whether AI systems can help make compromises across various use-cases \citep{conitzer_social_2024}.\footnote{In the Supplemental Information, we review a range of possibilities for formalizing a contractualist approach to value aggregation. See section ``Formalizing Contractualism.''}

%% file: sections/2_related_work.tex

\subsection*{Theoretical Foundations}

The literature on aggregating preferences spans rational decision-theory \citep{von_neumann_theory_1947, luce_games_1957, tversky_intransitivity_1969}, social choice theory \citep{sen_collective_2018}, and voting theory \citep{riker_theory_1968}. These theoretical frameworks offer distinct perspectives on how individual preferences can be consolidated into collective decisions.

Rational decision-theory, as the basis of understanding individual preferences, posits that individuals, when faced with multiple options, will choose the one that maximizes their utility \citep{von_neumann_theory_1947, tversky_intransitivity_1969,
kahneman_prospect_2013}.
Social choice theory, as an extension of rational decision-theory, analyzes individual preferences in a society and how they can be aggregated to reflect a collective preference \citep{sen_collective_2018}. It focuses on the design of mechanisms for making collective decisions, namely social welfare functions (SWFs). SWFs rank decisions based on their desirability to some group.\footnote{We exclusively look at cardinal SWFs: those which assume a numeric utility (outcome) for various groups. This stands in contrast to purely ordinal accounts, such as \citet{macaskill_normative_2016} introduces.}
We therefore use the framework provided by SWFs as our guide in this paper.  Another closely related line of work---voting theory---goes further to specifically addresses the methodology of preference aggregation in democratic decision-making processes, addressing concerns like strategic manipulation \citep{riker_theory_1968}; these issues are beyond the scope of the present work.



\paragraph{Aggregation Mechanisms}
There are many SWFs one might use to aggregate views.\footnote{Let \(A\) be the set of groups. Let \(B\) be a set of voting power (size) for each group in the space of \([0,1]^{|A|}\). Let \(C\) be the set of choices (or proposals). Let \(U\) in \(\mathbb{R}^{|B|\times|A|}\) for the cardinal case be the outcomes (utilities) associated for a particular group with a choice, where a particular choice, \(c\), and group \(a\), outcome is denoted \(u_a(c)\).} 
We will focus on two of the most popular. 
First consider the utilitarian SWF, e.g. as identified by \citet{von_neumann_theory_1947}, which we will term the ``Utilitarian Sum.'' Formally, this \textit{sums} the utility of available choices based on the amount of support for each. 

\begin{equation}
\argmax_{c \in C} \sum_{a \in A} u_a(c) \times b_a
\label{equation-sum}
\end{equation}

\noindent There are many ways in which the Utilitarian Sum is intuitively appealing.  For instance, it uses logic similar to what we use for dealing with \emph{empirical} uncertainty in a rational actor framework---simply do the action that leads to the best consequence taking into account how likely each consequence is and how good or bad it would be \citep{buchak_risk_2013}, equating degree of likelihood and belief.

The Utilitarian Sum also has important drawbacks. For instance, the Utilitarian Sum biases toward strong opinions of minority sub-groups---an issue called \textit{fanatacism} \citep{greaves_bargaining-theoretic_2023}.
It also makes \textit{inter-theoretic comparisons}: direct comparisons between the utilities of different groups which is  difficult to justify \textit{a priori}.
\citet{tarsney_vive_2021} critique various ways to make inter-theoretic comparisons, even though \citet{harsanyi_cardinal_1953} argues for their necessity.
The Utilitarian Sum has been widely studied for its use in preference aggregation (see below), particularly as it relates to empirical uncertainty \citep{greaves_bargaining-theoretic_2023}.

In contrast, \citet{kaneko_nash_1979} introduce the Nash Social Welfare Function which we will term the ``Nash Product.'' Formally, the solution to a Nash bargaining problem is 
to maximize the \textit{product} of utilities \citep{zamir_game_2013}:\footnote{The Nash Product is degenerate when utilities are less than one. We thus restrict ourselves to utilities of one or greater. This means that the outside option, or disagreement point, is also one.}

\begin{equation}
\argmax_{c \in C} \prod_{a \in A} u_a(c)^{b_a}
\label{equation-product}
\end{equation}





\noindent The Nash Product is more \textit{conservative} than fanatical. It trades-off maximizing aggregate benefit with capturing notions of fairness.

Another way to capture notions of fairness in a SWF is to use the Rawlsian lexical minimum, which maximizes the benefit to the least well off:

\begin{equation}
\label{eqn:leximin}
\argmax_{c \in C} \min_{a \in A} u_a(c) \times b_a
\end{equation}

Indeed, all three of equations \ref{equation-sum}, \ref{equation-product}, and \ref{eqn:leximin} are comparable. \citet{moulin_fair_2003} shows that a parameterized piece-wise function, where $\alpha$ tracks the degree of inequality aversion, results in the Nash Product when $\alpha = 1$, the Utilitariam Sum when $\alpha = 0$, and the lexical minimum when $\alpha = \infty$ \citep{bakker_fine-tuning_2022}:

\begin{equation}
\label{eqn:parameterized}
\argmax_{c \in C}
\begin{cases} 
    \sum_{a \in A} (u_a(c) \times b_a)^{1 - \alpha} &  0 \leq \alpha, \alpha \neq 1 \\
    \prod_{a \in A} u_a(c)^{b_a} & \alpha = 1  \\
\end{cases}
\end{equation}

\noindent As we discuss further below, while appealing for its privileging of egalitarianism, the Rawlsian minimum approach has important counter-intuitive implications \citep{harsanyi_can_1975, parfit_equality_1995, pena-rangel_political_2022, bruner_decisions_2018,frohlich_laboratory_1987}. The Nash Product has more inequality aversion than the Utilitarian sum (it is less fanatical) but not as much as the lexical minimum;
it exhibits diminishing marginal returns as utility increases linearly. \footnote{Indeed, the Nash Product is equivalent to the Utilitarian Sum under a log transformation of all outcomes: $\argmax_{c \in C} \sum_{a \in A} (\log u_a(c)) \times {b_a}$}  Our central experiments in this paper, therefore, compare intuitions about the Utilitarian Sum and the Nash Product.

These (and most other) SWFs assume that utilities are definable and known---and this carries non-trivial assumptions. For example, in economics, one might simply use a fungible price as a utility while utilities of outcomes in voting theory are not fungible. Furthermore, people may use different value functions to make decisions. In our experiments, we included both non-fungible and fungible quantities. \citet{mason_value_2023} reviews some theoretical concerns of such assumptions.\footnote{All of these SWFs can be set up to maximize a relative or absolute gain in utility. To do so, one simply changes the input utilities. In our case, we assume an absolute gain from zero.}


\subsection*{Normative Approaches}

How do we judge whether one aggregation mechanism is superior to another? 

\paragraph{Based on mathematical merits}
One approach examines the theoretical, mathematical trade-offs between SWFs, for instance by showing that in certain settings one SWF might not be mathematically optimal. 
There have been a number of such comparisons between the Nash Product and Utilitarian Sum-like approaches
\citep{rachmilevitch_nash_2023, rachmilevitch_egalitarianism_2019, rachmilevitch_nash_2015, thomson_nashs_1981}. More recent theoretical work on the Nash Product seeks to approximate it with mathematically analogous mechanisms \citep{mcglaughlin_improving_2020, branzei_nash_2017, ramezani_nash_2010}. \citet{kimbrough_meta-theory_2023} propose a number of game-theoretic heuristics (including the Nash Product) which people might use as a proxy to make moral choices.
The contrast between the Utilitarian Sum and the Nash Product also connects to recent debates in economics between (respectively) additive and multiplicative accounts of value, that is,  averaging via the arithmetic vs. the geometric mean \citep{peters_ergodicity_2019}.

\iftoggle{pnas}{
\figureScenarioPNAS
}{
\figureScenario
}

\paragraph{Based on intuition}

Another approach judges which aggregation mechanism  better matches the authors' intuitions.
Typically one examines isolated case-studies. For example, an author might claim that a SWF produces unintuitive results on a particular case study, using this as an argument for some other SWF.
Mathematicians, particularly decision theorists, exercise a degree of aesthetic judgement, or intuition, in defining the axioms of SWFs \cite{von_neumann_theory_1947, harsanyi_general_1988, harsanyi_cardinal_1955}. For example, \citet{luce_games_1957} introduce a number of classic cooperative games to gain intuition about game theory.

One prominent normative disagreement between contractualist and utilitarian mechanisms arose between \citet{rawls_theory_1971} arguing for a maximin account and \citet{harsanyi_can_1975} arguing for an expected value account. While both were operating under the assumption of a "veil of ignorance" style judgement,  each disagreed on the appropriate normative mechanism to use. 

Similarly, \citet{parfit_equality_1995} and \citet{nagel_equality_1995} both use specific (but different) imagined scenarios to justify opposing views on equality, as more recent work on equality does as well \citep{pena-rangel_political_2022}.



The use of authors' intuitions to make normative claims about value aggregation is common in other sub-areas of moral philosophy as well. For example, this issue is central to debates about ``moral uncertainty'', the puzzle of what to do if you believe different ethical theories to different extents \citep{lockhart_moral_2000, sepielli_what_2014, ross_rejecting_2006}. Different philosophers marshal their intuitions to argue that ethical theories can be aggregated either according to the consequentialist logic of the Utilitarian Sum \citep{macaskill_normative_2016, macaskill_why_2020} or a contractualist (agreement-based) logic \citep{newberry_parliamentary_2021}.\footnote{The bulk of MacAskill's argument comes in the form of specific scenarios which he uses to argue why intuition supports this favored mechanism. For example: "Julia works for a research funding body, and she has the final say over which of three proposals receives a major grant. ... The first, project A ... B, ... C ..." \citep{macaskill_normative_2016}. A similar strategy is used by \citet{newberry_parliamentary_2021}.} While \citet{newberry_parliamentary_2021} do not argue for the Nash Product in particular, they note that the Nash Product captures many of the virtues of their suggestion.\footnote{For instance, the Nash Product results in more equal outcomes, as per Equation~\ref{eqn:parameterized}. Nonetheless, \citet{greaves_bargaining-theoretic_2019} argue against the Nash Product in favor of the Utilitarian Sum, arguing against its \emph{conservatism}.}


\paragraph{Summing Up} The above approaches seek to justify one aggregation method over another based on theory, often using intuition to pick out single cases as intuitive counter-examples \citep{macaskill_normative_2016, newberry_parliamentary_2021, greaves_bargaining-theoretic_2023} or axiomatically seeking the most `rational' aggregation mechanism \citep{von_neumann_theory_1947, luce_games_1957, kaneko_nash_1979}. Less work has sought to ground the determination of the appropriate aggregation mechanism in studies of the decisions that people actually make.  We now turn to reviewing that literature.

\subsection*{The Empirical Approach}


\paragraph*{Behavioral Economics} When people make decisions between multiple outcomes, what approaches do they use? Questions like this are the domain of behavioral economics. Many works examine which resource distributions people favor, finding some evidence for a preference for equal allocations \citep{charness_understanding_2002, engelmann_inequality_2004, fehr_inequality_2006}.

Noting the fanatacism of the Utilitarian Sum, \citet{fehr_theory_1999, fehr_economics_2006} introduce a formalism sensitive to inequality (Equation~\ref{equation-inequality-aversion}). Subsequent work \citep{engelmann_inequality_2004, fehr_inequality_2006} has found support for an inequality aversion model over the Utilitarian Sum.

We consider such a model (an extension of \citet{fehr_theory_1999}) that directly modifies the Utilitarian Sum to be sensitive to the degree of inequality in outcomes:

\vspace{-\abovedisplayskip}
\begin{align*}
\argmax_{c \in C} 
& (1 - \alpha) ( \sum_{a \in A} u_a(c) \times b_a )   \stepcounter{equation}\tag{\theequation}\label{equation-inequality-aversion} \\
& -
\frac{ \alpha }{\binom{|A|}{2}} (\sum_{ a, a' \in A, a \neq a' } |u_a(c)
- u_{a'}(c)| )
\\
\end{align*}
\vspace{-3\belowdisplayskip}

\noindent The first term is just Equation~\ref{equation-sum} while the second term captures the amount of inequality across groups. $\alpha$ controls the degree of inequality aversion, with no aversion when $\alpha = 0$ and increasing aversion otherwise. 

%

Other work in behavioral economics focuses on the Nash Product, studying the effect of the disagreement point
\citep{bruce_impact_2012}, characterizing different bargaining strategies \citep{kroll_bargaining_2014}, and framing the Nash Product as a trade-off between utility or money \citep{berninghaus_trading_2008}.
In practice, \citet{yao_modeling_2021} find that in a certain modeling problem the Nash Product better fits the data than a Utilitarian approach, although they do not probe human intuitions directly.

\paragraph*{Empirical Philosophy}
Moral philosophers have increasingly used empirical inquiry to validate individual philosophers’ intuitions with the opinions of the crowd, making thought-experiments into real experiments,
 such as those about distributive justice \citep{frohlich_choosing_1992}.
\citet{bruner_decisions_2018}, for example, finds that when presented with a variety of scenarios of different resource distributions, participants prefer a strictly Utilitarian approach as compared to the Rawlsian minimum---participants maximize total utility not the utility for the least advantaged member (Equation \ref{eqn:leximin}). \citet{frohlich_laboratory_1987} present a similar result.

Similarly, \citet{bauer_need_2022}
study how various traits of agents change how much of a given resource participants distribute (though they do not focus on Utilitarian Sum or Nash Product in particular).

\paragraph*{Utilitarian Sum vs. Nash Product}

To our knowledge, the only study to empirically examine participants' responses regarding the Utilitarian Sum and Nash Product is \citet{binmore_focal_1993}. That paper investigates a variety of aggregation mechanisms, including the Nash Product and Utilitarian Sum, finding that it was more difficult to push participants to the Utilitarian Sum-supported answer. Building on their finding we ask a more direct question: which aggregation mechanism best accords with people's intuitions?



%% file: sections/3_method+results.tex
\section*{Studies}

Which method of aggregating preferences, of arriving at a compromise for a distribution of resources, is judged to be better---the Utilitarian Sum or the Nash Product?

\section*{Scenario generation}
\label{sec:scenario}
To study this, we generated scenarios where the Nash Product and the Utilitarian Sum disagree on the best way to aggregate value and designed an experiment with novel visual aids in which human and LLM participants judged which compromise was best (see Fig.~\ref{fig:scenario-charts}). Our paradigm also allows us to test the predictions of the Rawlsian Minimum (Equation~\ref{eqn:leximin}) and Inequality Sum (Equation~\ref{equation-inequality-aversion}). 

Unlike the focus on isolated cases of prior work,
we systematically canvas the space of possible simple value aggregation problems, testing a broad and representative set that distinguish the predictions of the Utilitarian Sum and the Nash Product mechanisms.

We asked subjects to imagine that their local health department was looking for feedback on how various proposals would affect the community. Then we asked participants to choose which of three proposals was the ``best compromise.'' Each scenario involved a different outcome for each of three groups across each of three proposals (Fig.~\ref{fig:scenario-charts}). These proposals would either decrease the average number of \textit{days to wait for an appointment}, decrease the number of \textit{minutes to travel for an appointment}, increase the \textit{years to live}, or decrease the \textit{cost of a medical visit}. We deliberately chose outcomes which were not always fungible monetary values in order to control for the effect of the kind of utility on the decision outcome. All scenarios were set up so that higher outcomes were more desirable, and thus the best outcomes---as predicted by the Utilitarian Sum or the Nash Product---\textit{maximized} these measures.


We generated two different sets of scenarios: Focused and Random.
For the ``Focused'' set, we generated scenarios that described three groups of constituents with equal bargaining power (i.e., number of people) deciding between three choices (proposals) whose outcomes laid in $\{1, 51, 101\}$.
Out of 19657 total scenarios meeting these conditions ($(3 ^3)^3$ minus duplicates), 162 (.8\%) resulted in a disagreement between the Utilitarian Sum and Nash Product. 
The ``Random'' set was constructed in the same way but included outcomes randomly sampled from the whole range $[1, 101]$ (e.g. $\{1, 17, 90\}, \{5, 53, 48\}, \ldots$). In this set, the Nash Product and Utilitarian Sum disagreed about 17\% of the time (see Table~\ref{tab:prevalence}).




Participants were presented with some scenarios where the Nash Product and the Utilitarian Sum disagree on which of the three policy choices would be superior as well as scenarios where the two aggregation mechanisms agree on which policy is best---the latter acting as a control condition.  






\figureMturkGptTest

\section*{Study 1: Human Participants}
\label{sec:human}

Each scenario (Fig.~\ref{fig:scenario-charts}) asked participants which of three proposals they thought was the "best compromise" between the groups. (See SI  \S~"Mturk Survey" for further details on experimental set-up.)


\tableMturk

\paragraph{Visual aids} Because of the numeric specificity of the proposals of the candidate aggregation mechanisms (as given in Equations~\ref{equation-sum}-\ref{equation-inequality-aversion}), our generated scenarios were necessarily quantitatively specific.  This allows for  precision in differentiating between the mechanisms, but comes with the challenge of overwhelming participants with numerical information.  We therefore developed a series of visual aids intended to assist participants in understanding the scenarios \citep{cleveland_graphical_1984}. This strategy has been fruitfully applied in prior psychological research.\citet{tversky_intransitivity_1969} showed pie-charts instead of ratios when asking about preferences.
\citet{eichler_different_2020, loibl_thinking_2024} show that certain visualizations improve participants' Bayesian reasoning ability.
We followed the visualization recommendations of \citet{shah_review_2002}.

We made two charts: 1) \textit{stacked bar charts} (Fig. \ref{fig:scenario-charts} C) showing the outcome on the y-axis with bars on the x-axis with width proportional to the normalized group size and 
2) \textit{3-dimensional bar charts} (Fig. \ref{fig:scenario-charts} B) displaying independent cuboids with length, width, and height proportional to the outcomes for each proposal for each group.\footnote{Interact with a demo of the visual aids here: \url{https://tinyurl.com/mu2h4wx4}.}
By making \textit{area} comparisons, the stacked bar charts visually correspond to the Utilitarian Sum--the proposal chosen by the Utilitarian Sum is the one which occupies the greatest area. By making \textit{volume} comparisons, the 3-dimensional bar charts visually correspond to the Nash Product--the proposal chosen by the Nash Product is the cuboid of maximal volume.

To check to see if there was an effect of chart type on participants' responses, we ran four different conditions: \texttt{No Charts}, \texttt{Both Charts} (ordered randomly on screen load time), \texttt{Area Chart} (the stacked bar chart), and \texttt{Volume Chart} (the 3-d chart).

\subsection*{Results}
\label{sec:human-results}

Throughout, we will focus on two different groups of scenarios: those in which the Utilitarian Sum and the Nash Product \textit{disagree} on which proposal is best (Disagreement Cases) and those in in which they \textit{agree} (Agreement Cases).
In the Agreement Cases, we report the percent of participants that select the response endorsed by both the Utilitarian Sum and the Nash Product. In the Disagreement Cases, we report the percent of participants that select the response endorsed by each of the Utilitarian Sum and Nash Product. (See the SI \S~"Study 1" for our survey qualification task.)


Our central finding is that in the Disagreement Cases, respondents overwhelmingly supported the Nash Product (see Fig. \ref{fig:test-conditions}A). In the Focused scenarios for all four conditions (\texttt{Area Charts}, \texttt{Volume Charts}, \texttt{Both Charts}, and \texttt{No Charts}) participants favored the Nash Product over random chance (Binomial test, $p < .001$) (see Fig. \ref{tab:results-mturk}A). We saw the same trend with the Random scenarios (See SI Fig. A.1-2 for details).\footnote{Pre-registered at \url{https://aspredicted.org/384_9FM}. Note that for simplicity, this study included only the \texttt{No Charts} condition.}


In the Agreement Cases, the majority of respondents across all conditions chose the correct answer (the answer both the Utilitarian Sum and the Nash Product agreed on; see Fig.~\ref{fig:mturk-control}A), confirming that participants understood the task and responded as expected to it. Notably, in the \texttt{No Charts} condition, performance dropped (though still was an improvement over chance), emphasizing the importance of the visual aids we provided. (Details in Fig. \ref{tab:results-mturk}A and Fig. \ref{fig:mturk-control}A for the Focused scenarios and SI Fig. A.1. for the Random scenarios.) 

\figureMturkLlmControl

\begin{figure}[t]
\includegraphics[width=\columnwidth]{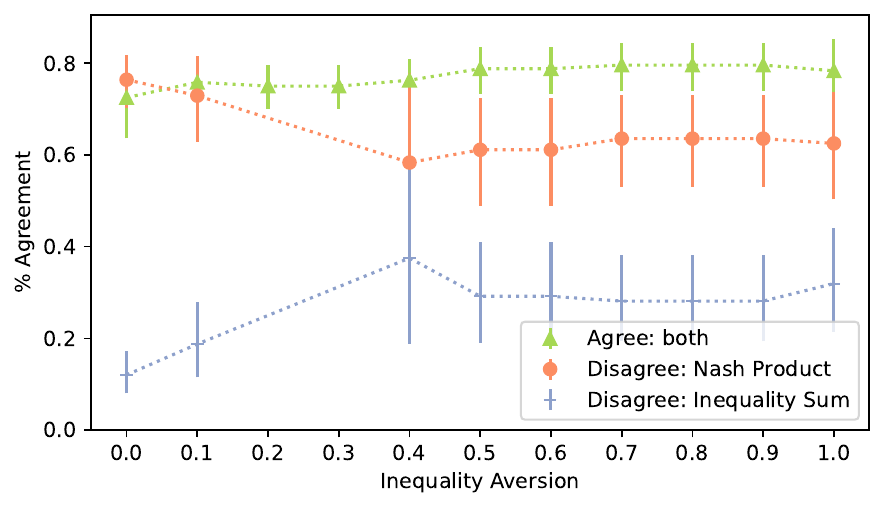}
\caption{The percent agreement between \textbf{human} participants and two aggregation mechanisms: the Nash Product (Eq.~\ref{equation-product}) and the Inequality Sum (Eq.~\ref{equation-inequality-aversion}, a variant of the Utilitarian Sum with a term to avoid inequality). The x-axis varies the inequality aversion parameter of the Inequality Sum: from no inequality aversion ($\alpha$=0; equivalent to the Utilitarian Sum) to only inequality aversion, ignoring aggregate utility ($\alpha$=1; similar to the Rawlsian Minimum, Eq. \ref{eqn:leximin}). The top, green line (\texttt{\^}) shows the proportion of participants who select the ``correct'' answer when the Nash Product and Inequality Sum agree on which proposal is best. The other lines track responses in the cases where the two mechanisms disagree. The middle, orange line (\texttt{o}) represents the proportion of subjects who endorsed the proposal consistent with the Nash Product and the bottom, blue, line (\texttt{+}) represents proportion of subjects who endorsed the proposals consistent with the Inequality Sum.  There were two points in which there were no disagreements between the Inequality Sum and the Nash Product, suggesting that the two mechanisms may be equivalent here.
(N=102 overlapping participants for each point.) Error bars show 95\% binomial confidence intervals. (See SI Fig. A.11 for the data of this plot.)}
\label{fig:inequality-aversion}
\end{figure}

We also compared the Nash Product with the Inequality Sum (Eq. \ref{equation-inequality-aversion}), a variant of the Utilitarian Sum with an added term to account for inequality aversion. The Inequality Sum can weight inequality aversion to different extents, depending on the setting of its free parameter ($\alpha$ in Eq. \ref{equation-inequality-aversion}). Fig.~\ref{fig:inequality-aversion} demonstrates that across the entire range of parameter settings for the Inequality Sum, participants strongly preferred the proposal suggested by the Nash Product when it disagreed with the Inequality Sum (Fig. \ref{fig:inequality-aversion}).

Finally, we compared predictions of the Nash Product with those of the Rawls Minimum (Eq. \ref{eqn:leximin}), and likewise find strong preference for the Nash Product (see SI Fig. A.3).

\section*{Study 2: LLM participants}
\label{sec:llm}

Large language models (LLMs) such as ChatGPT are already used for variety of human cognitive tasks \citep{ziems_can_2023} and, increasingly, in value aggregation tasks \citep{jiang_delphi_2021}. For example, \citet{bakker_fine-tuning_2022} directly use LLMs in an attempt to find agreement between different groups of people. Indeed, \citet{conitzer_social_2024} specifically argue that aggregation mechanisms like those we explore in this paper may be a better choice for the purposes of aligning AI systems.  Because of these trends, we sought to answer: \textit{Can any LLM serve as a model of preference aggregation?} Could LLMs be used as decision aides? To answer these questions, we replicated our human study with LLMs.

Given the current limitations of multimodal models \cite{yue_mmmu_2024}, we restricted our analysis to language-only models.  To as much as possible equate the experience of the LLM participants to that of our human participants (who were provided with visual aids in some conditions), we textually described the algorithmic steps of either the Nash Product (for the \texttt{volume chart} case), the Utilitarian Sum (for the \texttt{area chart} case), both, or neither (see SI Fig. A.13-14).

\subsection*{Results}
\label{sec:llm-results}

Here we report the results of the most performant model, \texttt{gpt-4-0613}. In the SI, we report the results of our experiments with \texttt{claude-3}, \texttt{claude-2.1}, \texttt{gpt-3.5}, and \texttt{davinci-002} (SI \S~"LLM Participants").


In the Disagreement Cases, like our human participants, \texttt{gpt-4} supported the Nash Product over the Utilitarian Sum and to a greater degree (Binomial test, $p < .001$; see Fig.~\ref{tab:results-mturk}B for Focused scenarios and SI Fig. A.1 for Random scenarios.) 
 
In the Agreement Scenarios, the performance of \texttt{gpt-4} diverged from our human participants. For the Focused scenarios, while \texttt{gpt-4} aligned with both the Nash Product and Utilitarian Sum more than chance in the \texttt{Volume Chart} condition ($p < .01$), \texttt{gpt-4} did not align with both more than chance in other conditions ($p's > .05$; see Fig. \ref{fig:mturk-control}B). This trend reversed for the Random scenarios in which \texttt{gpt-4} aligned with both the Nash Product and Utilitarian Sum more than chance in all conditions ($p < .001$; see SI Fig. A.1). \texttt{claude-3} showed similar performance to \texttt{gpt-4} but all smaller models showed even less alignment (see SI Fig. A.4-5).

\tablePrevalence 

\section*{Study 3: Prevalence}

How often do disagreements between the Utilitarian Sum and the Nash Product arise in real preference-aggregation problems? To answer, we analyzed three large and influential data sets for which this problem arises: Value Kaleidoscope \citep{sorensen_value_2023}, NLPositionality \citep{santy_nlpositionality_2023}, and Moral Machines \citep{awad_moral_2018}.

For example, the Value Kaleidoscope project \citep{sorensen_value_2023} aims to aid moral decision making. Type in a natural language dilemma, such as, "Telling a lie to protect a friend," and it outputs values that may support or oppose the dilemma, such as the "Duty to protect your friend's well-being" or the "Right to truthful information." Each of those values also assign a weight to each stance (e.g. 98\% supporting, 2\% opposing) as well as a relevance (e.g. 90\% relevant). This fits naturally into a value aggregation formulation we outline; both the Nash Product and Utilitarian Sum could be used to suggest whether one should support or oppose a given action once the values that support or oppose it (and their weights and relevance) are enumerated. In fact, \citet{sorensen_value_2023} explicitly rely on the Utilitarian Sum to do just this, (like many others before them) without considering the implications of this choice.  Using a large dataset of examples, we calculate how often disagreements over a final answer arise depending on whether the Nash Product or the Utilitarian Sum is used to aggregate value (see Fig. \ref{tab:prevalence}; for more details see SI \S~"methods" and SI Fig. A.12.) Even though disagreement scenarios at times occupy a small percentage of total scenarios, they can amount to a very large absolute number of decisions in the real world, especially as we increasingly see automated decision making systems deployed. Furthermore, up to now, the Utilitarian Sum has been the default mechanism for aggregating value (see SI \S~"Assumption of Utilitarian Sum"), although our work suggests that the Nash Product is more intuitive.


%% file: sections/5_discussion.tex
\section*{Discussion \& Conclusion}

When people aggregate values, what strategies do they think are best? In other words, \textit{which algorithm yields more intuitively plausible compromises, the Utilitarian Sum (an additive view) or the contractualist Nash Product (a multiplicative view)?} Our evidence shows that in cases in which the two mechanisms disagree, people overwhelmingly support the Nash Product, contrary to the current default assumption to use the Utilitarian Sum when values must be aggregated \citep{macaskill_normative_2016, greaves_bargaining-theoretic_2023, sorensen_value_2023, von_neumann_theory_1947}.

In the \texttt{No Chart} condition, when participants were presented with value aggregation problems involving raw numbers alone, they weakly favored the Nash Product over the Utilitarian Sum.  However, when provided with either an area-based or volume-based visual aid, their preference for the Nash Product became even more pronounced (Fig. \ref{fig:test-conditions}). This was particularly striking given that the visual aids were designed to represent (and thus bias toward) the calculations behind each of the aggregation mechanisms (the volume representation visualizing the Nash Product and the area representation visualizing the Utilitarian Sum).

Furthermore, in Agreement Cases, participants without a visual aid had weak or no significant alignment with both the Nash Product and Utilitarian Sum while participants significantly aligned with both mechanisms when provided a visual aid (Fig. \ref{fig:mturk-control}), underscoring the importance of the visual aids in getting clear, meaningful data to differentiate these views.

The Nash Product also appears to be preferred over other main algorithmic candidates for value aggregation, the Inequality Sum and the Rawlsian Minimum. In the case of the former, participants preferred the proposal favored by the Nash Product over that of the Inequality Sum for the full range of inequality-aversion parameters possible (Fig \ref{fig:inequality-aversion}).  The only parameterization where the Inequality Sum was equally as endorsed as the Nash Product was the one that produced no disagreements with the Nash Product.
Nonetheless, future work should explore additional ways of differentiating between the two mechanisms, for example by varying group sizes or by simply by focusing more narrowly on testing cases where the parameterizations of the Inequality Sum are most similar to the Nash Product.




Recently, scholars have begun to turn to contractualist accounts to explain the workings of the moral mind.  \citet{andre_evolutionary_2022} make an evolutionary argument that long-term concerns about an agent's social reputation explain the use of something like the Nash Product to ground and guide morality (see also, work by \citet{bruner_nash_2021}). 
\citet{levine_resource-rational_2023} argue, using a resource-rationality framework, that imagined approximations of a contractualist ideal (such as the one defined by the Nash Product) are pervasive in human moral thinking. Our findings corroborate these lines of work, providing empirical evidence that our participants have contractualist intuitions about the best way to solve value aggregation problems.  At the same time, however, we do not necessarily anticipate that participants are doing a complex multiplication problem in their heads to solve the value aggregation task we set in front of them.  It therefore remains an open question what algorithmic cognitive mechanisms allow participants to solve this task in line with the predictions of the Nash Product. (We explore a range of approaches in SI \S~"Formalizing Contractualism.")

As AI systems such as LLMs are increasingly deployed in value-laden decision making settings \citep{jiang_delphi_2021, conitzer_social_2024} and even to find compromises  \citep{bakker_fine-tuning_2022}, it is important to understand whether the aggregation mechanisms AI systems use align with the mechanisms people intuitively prefer. So: \textit{Can any LLM serve as a model of preference aggregation?} Performant LLMs such as \texttt{gpt-4} \textit{sometimes} display a similar preference to our human participants for the Nash Product over the Utilitarian Sum---they do model aspects of human preference aggregation. Indeed, models including \texttt{gpt-4} display systematically different biases in even slightly less constrained cases, calling into question their degree of alignment with human intuitions. Smaller and less capable models we studied diverged even farther from the behavior of our human participants, performing closer to chance across conditions (SI Fig. A.5). The performance of \texttt{gpt-4} suggests that more capable LLMs may be able to serve as cognitive models of value aggregation or used as compromise aides themselves, although further work should characterize in which domains performant LLMs are aligned with humans and in which they are not \citep{sucholutsky_getting_2023}.

%% file: sections/6_future_work.tex
\subsection*{Limitations \& Future Work}

Our studies compare one contractualist method of preference aggregation (the Nash Product) with the Utilitarian Sum (the canonical consequentialist method of aggregation).  However, contractualism comes in many forms and future work should explore whether formalizations of other contractualist mechanisms may capture people's intuitions better than the Nash Product or (perhaps most likely), whether different mechanisms capture intuitions in different circumstances.  Future work along these lines might aim to harness a parliamentary structure or the turn-taking nature of negotiation (perhaps using some sort of sequential decision making approach), which capture the spirit of contractualism and bargaining towards agreements.
(See SI \S~"Formalizing Contractualism" for a description of some attempts to do so.)

Our approach focuses on scenarios with fully-specified outcomes, group sizes, and a discrete number of available actions. In our prevalence analysis (Study 3), we show that these conditions are indeed sometimes met in real world cases that call for value aggregation.  However, the majority of cases where value aggregation is required will not have such information available.  A rich line of future work would involve explorations on how to weaken some of these constraints and assumptions. For example, systems might begin with natural-language scenarios and decompose into the formal models which we describe. Alternatively, one might attempt to replicate this work in an ordinal as opposed to cardinal setting.


Moreover, we surveyed only U.S.-based, adult participants and thus may have detected preferences that are constrained to that particular group.  Future work should explore individual and cultural differences in preference aggregation strategies to track the emergence and universality of the results we find here.

While in this paper we have focused on aggregating preferences between groups, the underlying formal mechanisms (but not necessarily the assumptions) are equivalent when aggregating between preferences within an individual. Consider: You sit down for dinner pining for a burger but torn up about animal welfare. What should you eat? In such cases, philosophers have asked what strategies a person should use when deciding between various normative theories  \citep{lockhart_moral_2000, macaskill_normative_2016, greaves_bargaining-theoretic_2023}, though quantitatively-precise investigations into this question from a descriptive perspective are just beginning \citep{levine_resource-rational_2023, maier_cheung_lieder_2023, Trujillo2024resource}. 
The methods we develop in this paper could be a fruitful approach to study what kind of mechanisms the mind uses to choose which moral mechanism to use when.


%% file: sections/7_conclusion.tex



\subsection*{Data Archival} All data and code to run these experiments are available on Github.  
\url{https://github.com/jlcmoore/valueaggregation}

\subsection*{Acknowledgements}

Thanks for the advice and comments from Xavier Roberts-Gaal, Bailey Flanigan, Fiery Cushman, Paul de Font-Reaulx, Max Kleiman-Weiner, Matthew Cashman, Toby Newberry, Hilary Greaves, Ronan Le Bras, Jena Hwang, anonymous reviewers in Stanford Class 329H, anonymous reviewers for AIES 2023, and the Mosaic team at AI2.
This work was funded in part by DARPA grant FA8650-23-C-7316 and ONR grant N00014-24-1-2207.

%% file: sections/8_appendix.tex
\subsection*{Aggregating Preferences in AI}
\label{app:sec:aggregating-preferences-ai}

Many subfields of AI, from game playing to computer vision, implicitly attempt to aggregate human preferences. Simply through next-word prediction, pre-trained language models encapsulate some preferences.



In a more general sense, there have been a variety of attempts to improve the moral reasoning ability of LLMs \citep{lourie_scruples_2021, jiang_delphi_2021}, sometimes paired with RL  \citep{hendrycks_what_2021, hendrycks_aligning_2021}. For example, \citet{pan_rewards_2023} test whether LLMs can avoid violating ethical norms in text-based adventure games, focusing on steerability. What these approaches lack is explicit adherence to a specific aggregation mechanism.

\paragraph*{Assumption of Utilitarian Sum}
Most existing attempts to deal with the problem of value aggregation in AI apply an algorithm in the family of Utilitarian Sum by making inter-theoretic comparisons or simply using the majority vote. This includes consequentialist approaches \citep{sorensen_value_2023, chen_just_2020, cousins_axiomatic_2021}, choice models \citep{martinho_computer_2021}, voting methods \citep{awad_moral_2018, noothigattu_voting-based_2018}, jury learning \citep{gordon_jury_2022}, and MDPs \citep{chaput_multi-agent_2021, loreggia_making_2022}.

\citet{feffer_moral_2023} critique such approaches by formally exploring what happens to a minority group if averaging methods (like the Utilitarian Sum) are implemented.
\citet{ethayarajh_authenticity_2022} further desribe how the assumptions of expected utility theory fail to work for collapsing the annotations of participnts.
These assumptions become even more pronounced when considering reinforcement learning from human feedback (RLHF) which explicitly optimizes models' adherence to humans' paired preferences \citep{lambert_history_2023}. These methods often assume human values are universal \citep{kirk_past_2023}.


\paragraph*{Other welfare functions}

Notably, \citet{takeshita_towards_2023} use the Utilitarian Sum to probe the responses of the Delphi \citep{jiang_delphi_2021} model, but they fail to compare against other game theoretic models and do not provide a systematic evaluation. \citet{sorensen_value_2023} can be seen as turning language-based moral dilemmas into the parameters of a bargaining game over moral dilemmas,
but they too end up using a form of the Utilitarian Sum.

\citet{bakker_fine-tuning_2022} train a reward model to rank individuals' agreement with the consensus-building statements of an LLM. They aggregate those preferences using three different social welfare functions: the Nash Product, Utilitarian, and Rawlsian. All three improve upon a model that does not incorporate individuals' preferences but \citet{bakker_fine-tuning_2022} find little differences between the SWFs. We see this as complimentary to our work; we focus explicitly on the Nash Product and the Utilitarian Sum, looking to find examples when the two theories come apart. 



\section*{Methods \& Results}

\subsection*{Scenario Generation}

We chose to generate games with groups of equal bargaining power (symmetric groups) because it is conceptually easier to grasp for participants and formally more concise for the Nash Product (which has an exponent in the asymmetric case). Note that our sampling strategy does not result in games favorable to either of the Utilitarian Sum or the Nash Product; it simply results in games when the two strategies disagree. Thus we expect that any bias toward one strategy in the resulting sample would hold across all disagreements. We assume that responses in the disagreement and non-disagreement cases are driven by similar principles, that the same logic drives the choice of whether to use the Utilitarian Sum or the Nash Product. As we later discuss with LLM responses, this assumption might not always hold.

Specifically, we generated a number of scenarios with different outcomes for three groups across each of three proposals. We randomly sampled 18 cases of disagreement between the Nash Product and the Utilitarian Sum from each set and 16 cases of agreement  for a total of 34 scenarios each.

We generated two sets of scenarios. The first we call our "Focused" scenarios with utilities lying in the set $\{1, 51, 101\}$. We chose linearly increasing utilities for the Focused set so as not to bias toward either the Nash Product or the Utilitarian Sum in terms of their underlying computation and, furthermore, we wanted utilities which laid on an understandable range to participants. The second set we call our "Random" scenarios with utilities randomly sampled from the set $\{1 \ldots 101\}$. We chose to test participants on two different sets of scenarios because we were concerned about artifacts from either set of scenarios overly influencing participants' choices. In the Focused set, the lack of variation in the utility values may have not been representative in sampling the space of all possible value aggregation problems. In the Random set, the difference between the utility values between any of the underlying scenarios in the set (e.g. one scenario with utilities $\{5, 59, 91\}$ and another with $\{40, 43, 87\}$) could too broadly sample the space of possible value aggregation problems and thus be less valid for comparison.

In five of the 16 cases of agreement in the Focused set of scenarios, there were two options that were rendered equally good by the Nash Product and the Utilitarian Sum.

\subsection*{Study 1: Human participants}

Our survey had four different scenarios in it for a total of eight questions including attention checks. We collected three participant responses for each unique survey. Each participant saw four different scenarios, one in each context:  \textit{days to wait}, \textit{visit cost}, \textit{minutes to travel}, and \textit{years to live} (as described in the main text) yielding 102 responses per condition when collapsing across scenarios.

We recruited participants through Mturk.
We used attention checks on each question and screened participants to only include those with a perfect score on a preliminary qualification task. This qualification required participants answer basic chart reading questions explained in the task.
(It appears in SI sec. "Qualification Task".)
19.94\% (646)
of 3239 respondents passed all 13 multiple choice qualification questions. All participants also had submitted at least 10k tasks on Mturk, were living in the United States, and had a task approval rate of greater than 97\%. The average response time across all qualifications was 10.6 minutes (STD 7.9).
Having paid \$3 (USD) per qualification task,
this averages to \$17.0 an hour.
We only allowed each qualified participant to submit one survey across all conditions. On average, a submission took 6.2 minutes (STD 3.5) 
and we paid \$3 per submission, yielding an average hourly wage of \$29.

Of those who passed our qualification task and went on to complete the main experiment, 15\% of respondents failed at least one attention check.
We excluded these respondents from our analysis and collected more responses to replace theirs until we had 100\% coverage of all scenarios with contexts. Note that we had three different participants respond to exactly the same set of four scenarios with attached contexts. Importantly, we compare the aggregated \textit{scenarios} (with about 14.8 average responses each) not the scenarios with added context.

In tie cases, one of the Utilitarian Sum or the Nash Product tied between proposals and hence the two mechanisms do not fully agree. As expected for these uncertain cases, across conditions we saw a lower median agreement rate across respondents (SI Fig. \ref{app:mturk-max-ties-right}).

\subsubsection*{Statistical Tests}

To calculate the significance values in (main text) Figures 2 and 3 as well as \ref{app:gpt4-scenarios-101-outcomes}, \ref{fig:test-conditions-2}, \ref{app:rawls-nash}, \ref{app:gpt4-scenarios-101-outcomes-control}, and  \ref{app:fig:llm-control}, we used a binomial test using the python \texttt{scipy} package. In the Agreement cases, we coded each subject's response as a 1 (a ``success'' in the language of binomial tests) if they agreed with the proposal suggested by the Nash Product and the Utilitarian Sum (or the Inequality Sum) and a 0 otherwise. In the Disagreement cases, we coded each subject's response as a 1 if they agreed with the proposal from each respective aggregation mechanism (Nash Product, Utilitarian Sum, Inequality Sum). In all cases we assume a null hypothesis of random guessing which, given that we asked three questions, was always 33\%. We passed these parameters to the relevant binomial functions to calculate the p-value as well as the 95\% confidence intervals.

In the case of the comparison between the Nash Product and the Inequality Sum, Figure 5 in the main paper, the CIs vary because the number of disagreeing scenarios varies. There were no disagreements between the Nash Product and Inequality Sum at an inequality aversion of $.2$ and $.3$. There, the data covers just the Focused scenarios and the \texttt{Area Charts} condition. See SI \ref{tab:inequality-aversion}.

\subsection*{Study 2: LLM Participants}

We report experiments on a number of large closed-source models from OpenAI (\texttt{gpt-4-0613}, \texttt{gpt-3.5-turbo-16k-0613}, \texttt{davinci-002}) and Anthropic (\texttt{claude-2.1}, \texttt{claude-3-opus-20240229}).

\subsubsection*{Methods}
We prompted models with the answers to a few qualification task questions (quasi-few-shot), including the textual versions of the volume and area charts.
(See SI Fig. \ref{app:claude-example-area}.)
We say quasi-few-shot because the qualification tasks had no mention of "compromise".
These examples we provided LLMs were made in a chain-of-thought (COT) style, beginning with "Let's think step by step"
\citep{wei_chain_2022}. (Examples of our prompts appear in SI Fig. \ref{app:claude-example-area} and \ref{app:claude-example-volume}.)
To better understand the distribution of model responses, we tested at a temperature of 1 and took 10 samples for each query, turning the answers into a distribution of responses.


Having defined a multiple-choice question answering task, we follow \citet{fu_chain--thought_2023} in prompting models to summarize their (often verbose) responses in a single letter (A, B, etc.). While smaller models might struggle to respond in such a paradigm despite containing relevant knowledge \citep{hu_prompt-based_2023, burns_discovering_2022}
we found no such issue in the case of the large models on which we tested.
For those models which gave API access to log probabilities, we follow \citet[app. 3]{santurkar_whose_2023} in gathering a distribution over model responses.

In addition to running the main scenarios as we did with our human participants, we wanted to test if LLMs were capable of performing the underlying calculations of each aggregation mechanism---could they do the math of equations 2 and 4? We did so by administering a version of the qualification task we used to screen human participants in the chart conditions, asking models to choose the proposal with either the largest \textit{volume} (Nash product) or \textit{area} (Utilitarian Sum). Here we prompted models with questions without any preceding context or examples (0-shot). When prompted to choose the proposal of largest \textit{volume} or \textit{area} (instead of the "best compromise"), we found that models agreed with the Nash Product or the Utilitarian Sum both in agreement (SI Fig. \ref{app:llm-qual-control}) and in disagreement scenarios (SI Fig. \ref{app:llm-qual-test}).
In the qualification task, when we prompted models to answer which option yielded the greatest "volume" (for the Nash Product) or "area" (for the Utilitarian Sum) we found that all models except \texttt{davinci-003} (which performed at chance) performed quite well (agreed with the Nash Product or the Utilitarian Sum, respectively), both in agreement and in disagreement cases.
For example, investigating the step-by-step math of the models demonstrates many mistakes (e.g. with exponentiation and multiplication, see SI Fig. \ref{fig:claude-qualification}).

\subsubsection*{Results}

In the agreement scenarios, all models had lower mean alignment rates with the Nash Product and the Utilitiarian Sum than \texttt{gpt-4} and \texttt{claude-3}, across conditions (SI Fig. \ref{app:fig:llm-control}), regardless of the decision aid they were shown---whether they were shown nothing in addition to the scenario (\texttt{none}), the textual description of the Utilitarian Sum (\texttt{area}), or the description of the Nash Product (\texttt{volume}) (see SI Tab. \ref{app:results-llm}).
All models achieved a lower mean agreement when not shown the descriptions as compared to when shown the descriptions. Across conditions, \texttt{gpt-3.5} performs much worse than in the qualification task, despite the fact that simply applying the Utilitarian Sum (which it can do) would have sufficed.
Similarly to our human subjects, models were less consistent on the agreement tie cases (SI Fig. \ref{app:llm-max-tie-right}).

In the disagreement cases, we saw a similar trend as in the human experiment in which the performant models (all but \texttt{davinci-002}, which performed at chance) overwhelmingly achieved a higher rate of alignment with the Nash Product than with the Utilitarian Sum (Fig. 2 and SI Fig. \ref{fig:test-conditions-2}). 

\textit{Are LLMs even able to compute the Nash Product and the Utilitarian Sum?} Some are. When asked to choose the proposal of greatest "area" or "volume", instead of the "best compromise", \texttt{gpt-4} successfully mirrored the calculations of the Utilitarian Sum and the Nash Product, respectively, performing significantly better than chance (see Figs.~\ref{app:llm-qual-control},~\ref{app:llm-qual-test}, where we also report results of the performance of other LLMs). Therefore, a lack of performance on the "best compromise" task cannot be due to the fact that an LLM is inherently unable to perform the necessary calculations; an LLM simply does not use that mechanism at the appropriate time. As we report in the main text (see Fig. 4B), \texttt{gpt-4} did not perform better than chance at choosing the best compromise, except in the "Volume" condition. (Other models show similar or worse results \ref{app:fig:llm-control}.)  This suggests that even when LLMs are able to do the calculations necessary to be aligned with human responses, they do not know when to apply the necessary calculations in problems of value aggregation and thus,  at least in these scenarios, fail to capture the intuitions of human participants.



\subsection*{Prevalence}

The Value Kaleidoscope project \citep{sorensen_value_2023} uses the Commonsense Norm bank data set  \citep{jiang_delphi_2021} as a way to train a model to generate unique values, their relevance, and their degree of support for any free-form natural language input. For example given the statement "If Owen's friend Jake is having trouble with his girlfriend, and Owen wants to help Jake feel better, is Owen sending flowers to Jake's girlfriend in Jake's name, hoping to patch things up moral?", Kaleido generates values such as "Duty to protect Jake's emotional well-being" which supports the statement and "Jake's girlfriend's right to not be deceived" which opposes the statement. We input those values (e.g., SI Tab. \ref{app:naturalistic-values}) into the different aggregation mechanisms assuming relevance corresponds to voting power and the degrees of support (the values of support, oppose, and either) as the outcomes for each group (value). In the Jake example, it turns out that under these assumptions the Utilitarian Sum opposes the action and the Nash Product supports it.

The NLPositionality data sets \citep{santy_nlpositionality_2023} gather scores from "It's very good" to "It's very bad" from crowd workers from different demographic groups on the acceptibility of various statements. We look at just their SocialChemistry split. We use these scores as utilities and the choices as actions treating different subsets of demographic groups as the groups and then set the voting power as the normalized size of that sub group. For example, when comparing the demographic `20-30' and `10-20' for the statement "children doing risky things", the Utilitarian Sum chooses "It's okay" while the Nash Product chooses "It's bad" (SI Tab. \ref{app:naturalistic-values}).

The Moral Machine project \citep{awad_moral_2018} gathered  millions of responses worldwide about whom an autonomous vehicle should save in various trolley-car-like dilemmas with changing categories. For the utilities, we used their aggregated AMCE scores which show the preference for each country for one category over another. For the voting power, we use UN population data to estimate the belief for different country preferences reported \citep{united_nations_department_of_economic_and_social_affairs_population_division_world_2022}.
Because of resource limits we compared five countries between each other at a time and counted the disagreements across various attributes. For example, when comparing Italy, Colombia, the UAE, Panama and Slovenia based on the values (SI Tab. \ref{app:naturalistic-values}), the Utilitarian Sum favors saving more characters overall while the Nash Product favors favoring humans over other species.


\input{sections/formalizing_contractualism}


%% file: sections/formalizing_contractualism.tex
\section*{Formalizing Contractualism}
\label{app:sec:formalizing}

What is the best way to aggregate value?  Below we survey a range of algorithmic implementations of \emph{contractualist} (agreement-based or negotiation-based) answers to this question.  


\paragraph{Nash Product}

The Nash Product has long been used to model bargaining; it was initially introduced as the Nash Bargaining Solution \citep{nash1950bargaining, kaneko_nash_1979}. It allows for comparison between different choices available to multiple parties wherein each party to the bargain also has the option to choose not to bargain and instead choose what is called the disagreement point (the status quo or outside option). No verbal negotiation goes on between the parties. The Nash Bargaining Solution is simply the choice which maximizes the product of each party's gains over the status quo--the choice which maximizes cooperation between the parties to the negotiation.
For this reason, following other scholars, we contend that the Nash Product provides a \emph{contractualist} \citep{andre_evolutionary_2022, levine_resource-rational_2023} account of value aggregation--one built around agreement. 

This is in contrast to the dominant consequentialist approach of the Utilitarian Sum. Indeed, we began this work as an attempt to question some of the assumptions that the Utilitarian Sum makes, namely that it engages in \emph{intertheoretic comparisons}, it equates individuals utilities, and it is prone to \emph{fanaticism}, it can be swayed by strong opinions of minority groups. The Nash Product is not as susceptible to fanaticism as the Utilitarian sum but it fundamentally makes intertheoretic comparisons on the Pareto frontier.\footnote{The Nash Product itself applies a structural normalization over the input utility values while the Utilitarian Sum has to be supplemented with one--usually the variance \citep{greaves_bargaining-theoretic_2023}.} Furthermore, as noted above, the Nash Product formally requires the specification of a disagreement point, or outside option \citep{greaves_bargaining-theoretic_2023}. Often the Nash Product is used on utilities greater than or equal to one (lest the product become infinitesimal) and so requires a structural transformation to a different range, usually, e.g. $[1, \infty)$---a similar structural transformation as is suggested for the Utilitarian Sum. 

The Nash Product depends on the utility \textit{gains} in a way that Utilitarian Sum does not. Thus what counts as a gain is contingent on what each agent's outside option or disagreement point is.
The disagreement point is what happens if no majority is reached--often either a utility of zero, some extreme value, or the outcomes of some other default strategy.
Define the disagreement point, \(\mathbf{d} \in \mathbb{R}^{|A|}\) such that the outcome of the disagreement point is also an available utility for each agent, \(U \cup \{ \mathbf{d} \}\) \citep{greaves_bargaining-theoretic_2023}.
Still, we do not find the specification of a disagreement point as a significant assumption. 
How often is it the case that a decision has specified all of the utilities for the potential proposals or actions but does not have a specified disagreement point? Fundamentally, assessing the utilities of actions is not that different from assigning utilities for a disagreement point (a sort of null action).

Nonetheless, it is possible to circumvent this issue by stipulating utilities at the disagreement point (or stipulating the change in utilities from the disagreement point for every action available in the set). This is what we do in our studies.

Still, there are a variety of other formal approaches one might take to contractualism. 


\paragraph{Turn-Taking Games} One approach is to model a bargain as an extensive, turn-taking game like chess. This has the benefit of avoiding any intertheoretic comparison: each group imagines their best choice given the choice of every other group in which groups have different voting power--similar to a parliament. In order to encourage coalitions in such a game, \citet{newberry_parliamentary_2021} suggest setting the utility of a choice in proportion to the weights each vote receives (groups by group weight) but then choosing the best option by majority vote. They call their approach "proportional chances." For a two player game assume some voting mechanism (social
welfare function), \(F\), which operates over the outcomes, $U$, group beliefs, $B$, and choices, $C$, where $c_i \in \{0, 1\}$, $1$ if group $i$ chose that choice and $0$ otherwise and $\mathbf{u}(c)$ is the vector of $U$ for choice $c$. $F_{pc}$ is the function for proportional chances.

\begin{equation}
\max_{c \in C} \max_{c' \in C}{ F(U, B, \{c, c'\})}
\end{equation}

\begin{equation}
F_{pc} = \max_{c \in C} \left (\mathbf{u}(c) \sum_{a \in A} c_a \times b_a \right)
\end{equation}

What becomes apparent is that taking the proportion is not strictly necessary for each player to incorporate the others' actions. It can also cause free-riding. 
Consider an example which we have set up to appear like an intuitive opportunity for negotiation to occur. A plurality group, "a" has the highest voting power and prefers an option much dispreferred by the two minority groups. Each minority group, "b" and "c" prefers an option dispreferred by the rest, "2" and "3" respectively.  The minority groups want choice "4"  second-best. They should collaborate to vote for this option. All of the terms in $b_{a} > b_{b}, b_{c}$, are greater than zero, and $u_{c, b}(4) < u_{c}(3),  u_{b}(2)$.

Nevertheless, when cast as a proportional chances game, no cooperation emerges here because either of the minority groups can free ride off of the others' vote for the second-best option and still vote for their preferred option (at least as they see it in the game tree).
For example, consider whether "b" chooses to vote for "2" or "4" given that "a" votes for "1" and "c" attempts to bargain by voting for "4"; the utility of the former will always dominate the utility of the latter. 

\begin{align*}
b_{a} u_{a}(1) + b_{b} u_{b}(2) + b_{c} u_{c}(4) & > b_{a} u_{a}(1) + b_{b} u_{b}(4) + b_{c} u_{c}(4) \\
b_{b} u_{b}(2) & >  b_{b} u_{b}(4) + b_{c} \\
\end{align*}

Still, many other voting mechanisms, $F$, might be used. If the strict majority vote is used, it will fail to give answers when only a plurality is reached; it will not be complete. Instead, terminal utilities can simply be the players' respective outcomes for what would happen if each player voted a certain way, using the weighted majority vote. Call this approach the maximax disagreement (mmd), $F_{mmd}$

\begin{equation}
F_{mmd} = 
\begin{cases} 
\mathbf{0} & \max_{c \in C} \sum_{a \in A} c_a \times b_a < .5 \\
\mathbf{u}\left ( \sum_{a \in A} c_a \times b_a  \right ) & otherwise
\end{cases}
\end{equation}

Unfortunately, turn-taking games are prone to dominant strategies by the first player. Depending on the social welfare function used it can become an ultimatum game (the player to go
first dictates the outcome) or yield different solutions based on which
agent chooses first.



For example, consider a game with two groups, $a$ and $b$, of equal bargaining power considering three choices, "a-pref", "bargain", and "b-pref", where $u_{a}(a\mbox{-}pref) \succ u_{a}(bargain) \succ  u_{a}(b\mbox{-}pref)$ and $u_{b}(b\mbox{-}pref) \succ u_{b}(bargain) \succ u_{b}(a\mbox{-}pref)$. In this case, the outcome of any turn taking game always depends on which group votes first in the game tree and the groups will never choose the bargain option.

\paragraph{Strategic Games}
More promising would be a strategic, non turn-taking, equilibrium selection approach \citep{harsanyi_general_1988}. Unfortunately, these are notoriously complicated and case specific. 
For example, neither of the outcome (utility) vectors $(1, 10, 100)$ nor $(100, 100, 1)$ Pareto-dominates the other. Nonetheless, it seems obvious that the second is  preferred. What about $(1, 51, 10)$ compared to $(1, 10, 51)$ or $(1, 51, 10, 10)$ compared to $(9, 2, 52, 9)$? These issues are legion.

\paragraph{Shapley Values}
Coalition-forming approaches such as Shapley values (which still make intertheoretic comparisons) are also worth exploring.
In such coalition-forming
games, groups with asymmetric bargaining power form
coalitions with each other, each coalition perhaps in favor of a certain
choice. Allow some function to describe which coalition is
successful, usually a loose majority vote. Here the difficulty is how to
assign credit to each of the individual groups in a coalition. The
standard interpretation describes the dispersal of
some fixed, usually monetary, quantity between agents. The Shapley value
is one approach to give the most credit back to the agent who most
contributed to the success of the particular coalition. This may not be
tenable unless intertheoretic comparisons are allowed. Other
interpretations are possible and should be explored in the case of moral
negotiation, perhaps as a kind of voting credit in a sequential game,
capturing the sense of "you helped me out last time" (similar to the approach used in
\cite{ecoffet_reinforcement_2021}).

\paragraph{Summing Up}

Because of the non-completeness of turn-taking games, the multiple equilibria of strategic games, and the lack of clarity on how to resolve credit assignment with Shapley values, we chose to use the Nash Product to model contractualist reasoning in this study: it is complete and is the standard choice for modeling bargaining.

Still, it may simply be that no game theoretic approach sufficiently captures the variance of human negotiation. In that case, language-based approaches might be the best way forward, e.g. if we could accurately simulate different perspectives in various LLMs and literally put them in conversation with each other. We leave such an endeavor for future work.

%% file: sections/appendix_figures.tex

\begin{figure}[tbhp]
\centering
\includegraphics[width=0.2\textwidth]{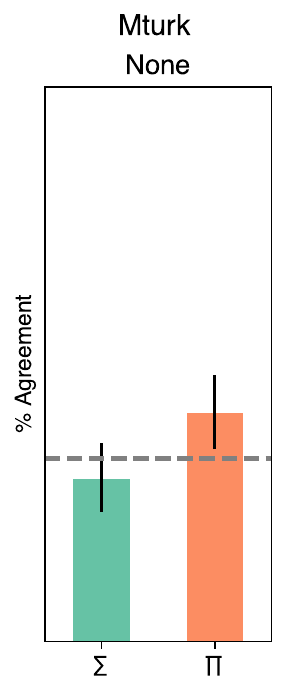}
\includegraphics[width=0.4\textwidth]{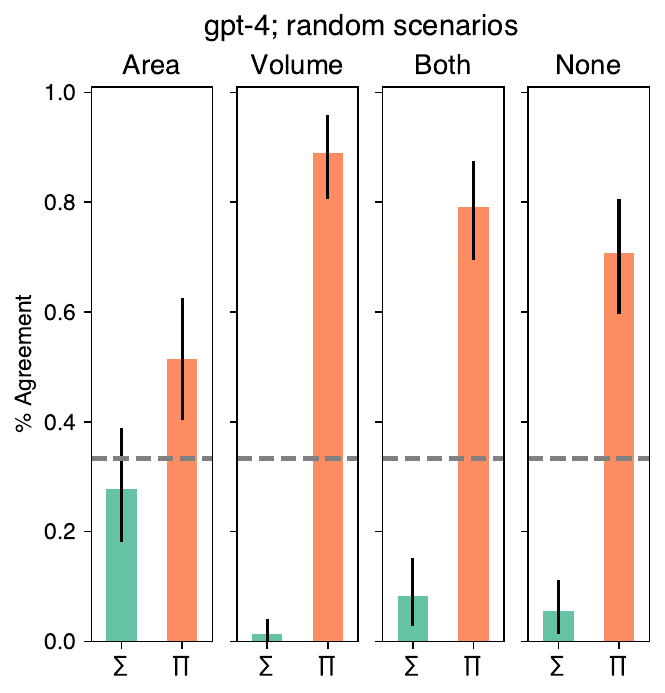}
\caption{Results for the Disagreement Scenarios.
Responses to the ``Random'' set of scenarios in which outcomes ranged from $[1 \ldots 101]$ instead of $\{1, 51, 101\}$, as shown in the main text (see Fig. 2).
The percent agreement of human (Mturk) participants and \texttt{gpt-4} with two different value aggregation algorithms: the Utilitarian Sum (an additive model, shown in green with the $\Sigma$ symbol) and the Nash Product (a multiplicative model, shown in orange with the $\Pi$ symbol).  (N=102 per condition.) The panels represent the different visual aids that participants received: \texttt{area}, \texttt{volume}, \texttt{both}, and \texttt{none}. The dashed line at 33\% indicates random guessing. (Participants always selected from three options.) Error bars show 95\% binomial confidence intervals.}
\label{app:gpt4-scenarios-101-outcomes}
\end{figure}

\begin{figure}[tbhp]
\centering
\includegraphics[width=.2\textwidth]{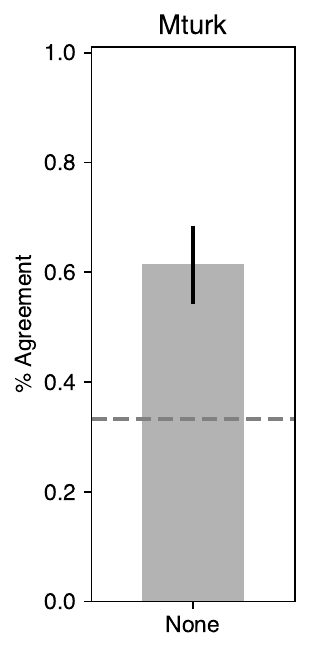}
\includegraphics[width=0.4\textwidth]{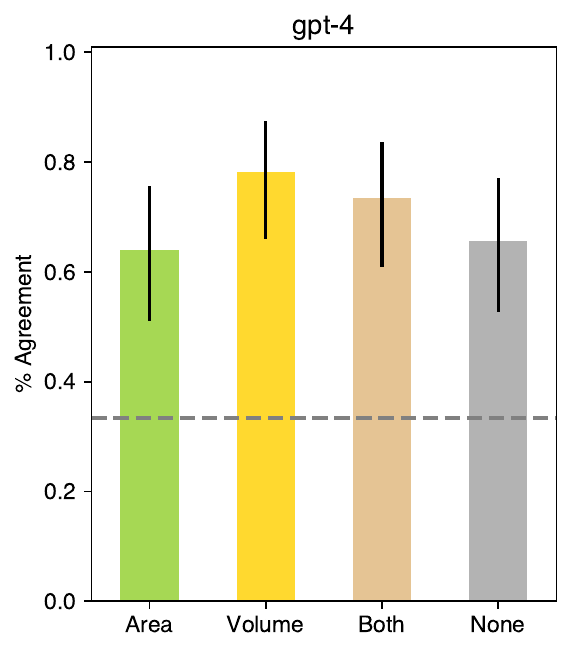}
\caption{Results for the Agreement Scenarios.
Responses to the ``Random'' set of scenarios in which outcomes ranged from $[1 \ldots 101]$ instead of $\{1, 51, 101\}$, as shown in the main text (see Fig. 2).
The percent agreement of {human} participants (Mturk) and {\texttt{gpt-4}} with the Utilitarian Sum and the Nash Product. (N=102 per condition.) The panels represent the visual aids participants received: \texttt{area}, \texttt{volume}, \texttt{both}, and \texttt{none}. The dashed line at 33\% indicates random guessing. \\[.25\baselineskip]
High agreement with the Utilitarian Sum and the Nash Product when both agree indicates that the two capture what participants intuit by a "best compromise."}

\label{app:gpt4-scenarios-101-outcomes-control}
\end{figure}

\begin{figure}[tbhp]
\centering
\includegraphics[width=.5\textwidth]{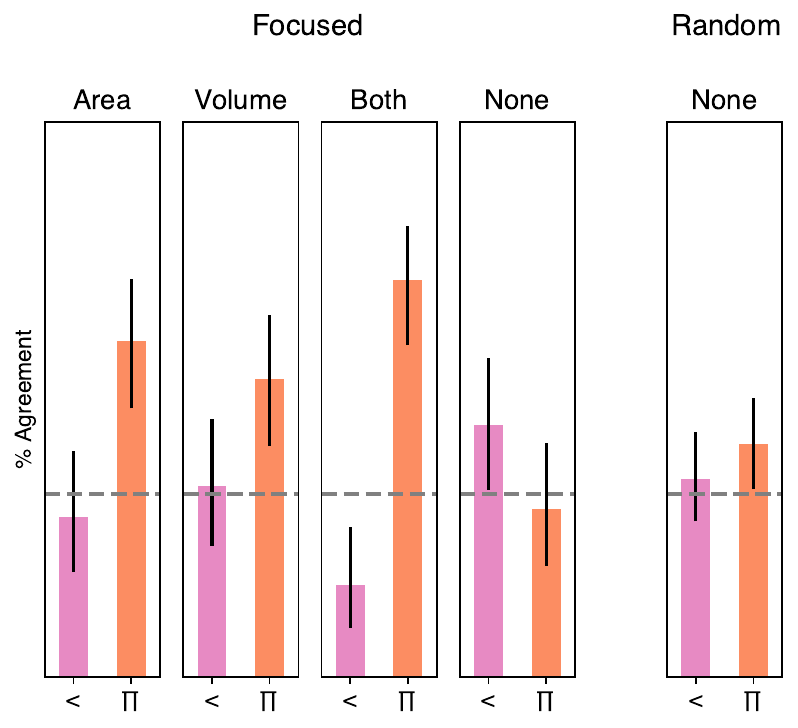}

\caption{
The percent agreement of human (Mturk) participants and \texttt{gpt-4} with two different value aggregation algorithms: the Rawlsian Lexical Minumum (shown in magenta with the $<$ symbol) and the Nash Product (shown in orange with the $\Pi$ symbol) on cases in which the two mechanisms disagree.  (N=102 per condition.) The panels represent the different visual aids that participants received: \texttt{area}, \texttt{volume}, \texttt{both}, and \texttt{none}. Data on both the ``Focused`` and the ``Random'' set of scenarios in which outcomes laid $\{1, 51, 101\}$ and $[1 \ldots 101]$, respectively.
The dashed line at 33\% indicates random guessing. (Participants always selected from three options.) Error bars show 95\% binomial confidence intervals.}

\label{app:rawls-nash}
\end{figure}

\begin{figure*}
\centering
{
\begin{minipage}[t]{.24\textwidth}
\centering
\includegraphics[width=1\textwidth]{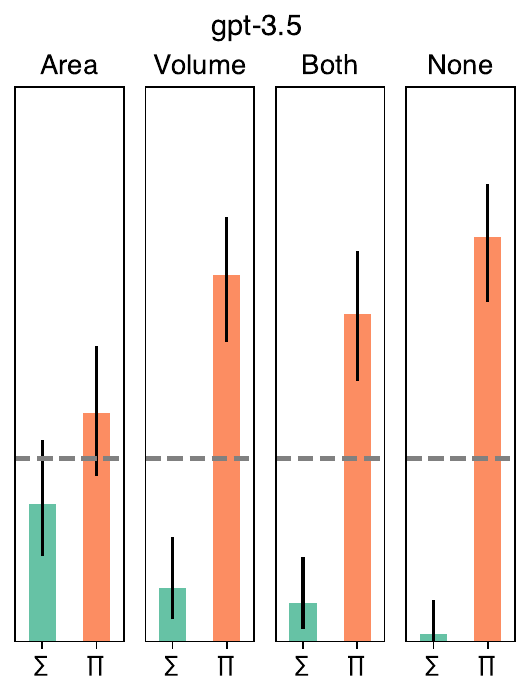}
\end{minipage}
\begin{minipage}[t]{.24\textwidth}
\centering
\includegraphics[width=1\textwidth]{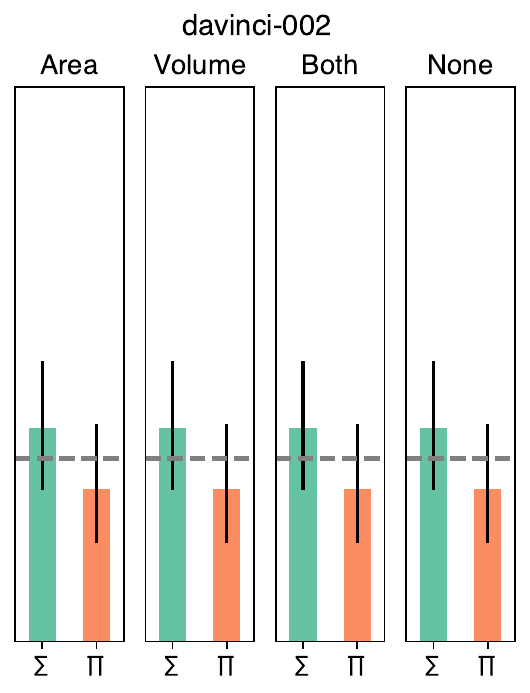}
\end{minipage}
\begin{minipage}[t]{.24\textwidth}
\centering
\includegraphics[width=1\textwidth]{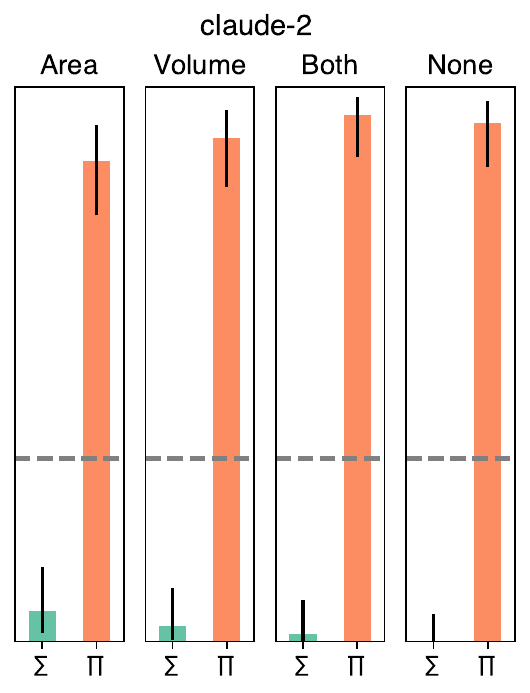}
\end{minipage}
\begin{minipage}[t]{.24\textwidth}
\centering
\includegraphics[width=1\textwidth]{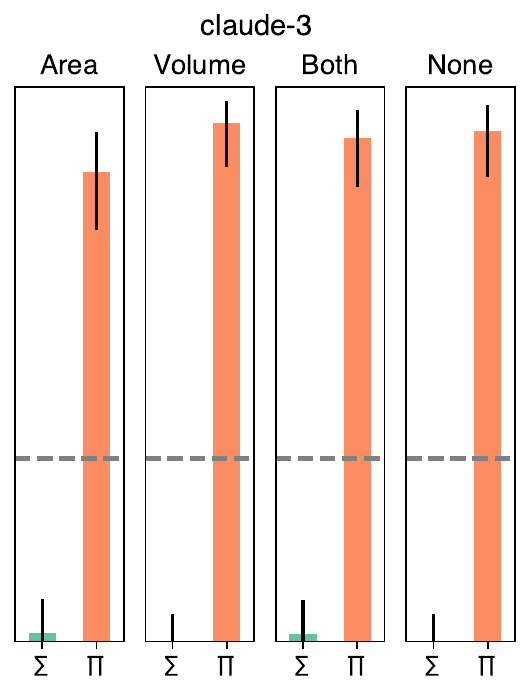}
\end{minipage}
}

\caption{Additional Large Language Model (LLM) responses to the Disagreement Scenarios (compare to GPT-4 responses in Fig. 2B in the main text.) Figure shows the percent models are aligned with two different value aggregation algorithms: the Utilitarian Sum (an additive model, shown in green with the $\Sigma$ symbol) and the Nash Product (a multiplicative model, shown in orange with the $\Pi$ symbol) on cases in which the two mechanisms disagree. The panels represent the different visual aids that models received (described textually): \texttt{area}, \texttt{volume}, \texttt{both}, and \texttt{none}. The dashed line at 33\% indicates random guessing. (Models always selected from three options.) Error bars show 95\% binomial confidence intervals. (See Fig. \ref{app:results-llm} for means and significance values.)}
\label{fig:test-conditions-2}
\end{figure*}

\begin{figure*}
\includegraphics[width=1\textwidth]{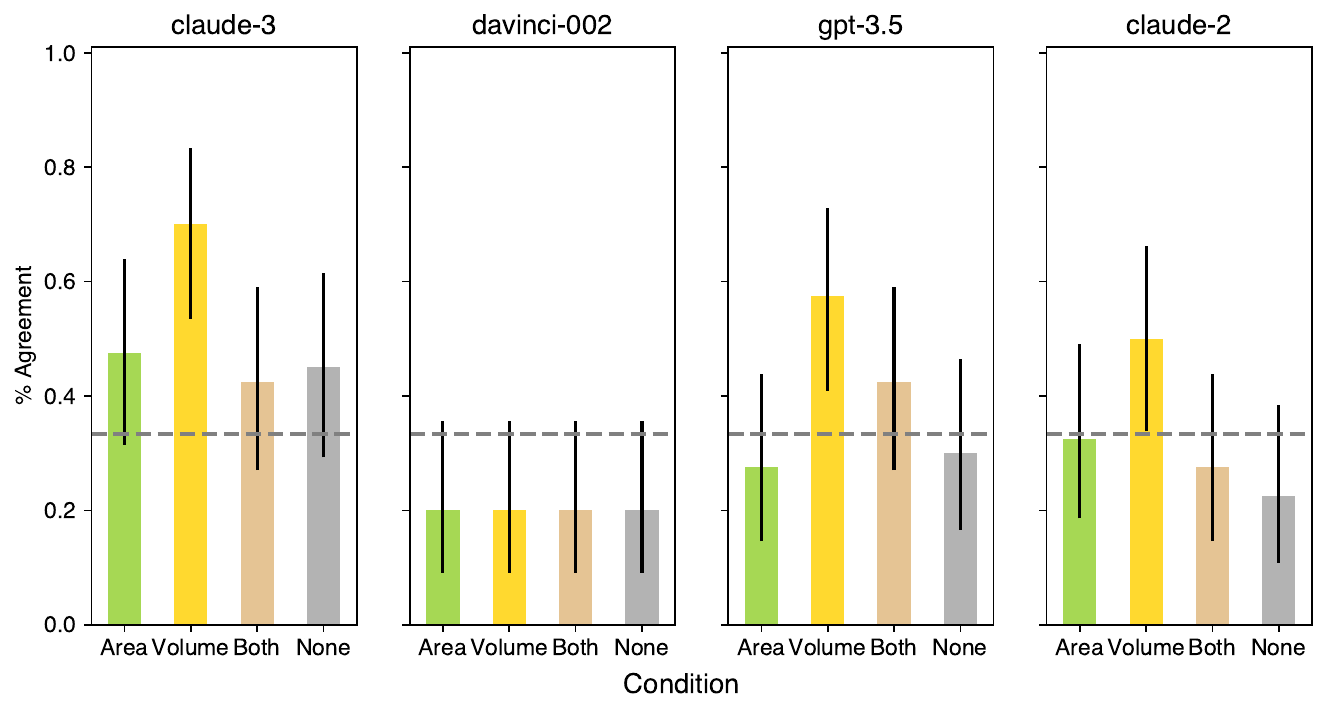}
\caption{Additional Large Language Model (LLM) responses to the Agreement Scenarios (compare to GPT-4 responses in Fig. 4B in the main text.) Figure shows the percent models are aligned with the Utilitarian Sum and the Nash Product on each of the agreement scenarios across the \texttt{area}, \texttt{volume}, \texttt{both}, and \texttt{none} conditions. In comparison to the results from human participants, the lower agreement of LLMs (except \texttt{gpt-4}) with the Utilitarian Sum and the Nash Product when both agree indicates that computations besides the Utilitarian Sum and the Nash Product drive the choice of a "best compromise." (See Fig. \ref{app:results-llm} for means and significance values.)}
\label{app:fig:llm-control}
\end{figure*}


\begin{figure}[tbhp]
\centering
\includegraphics[width=0.5\textwidth]{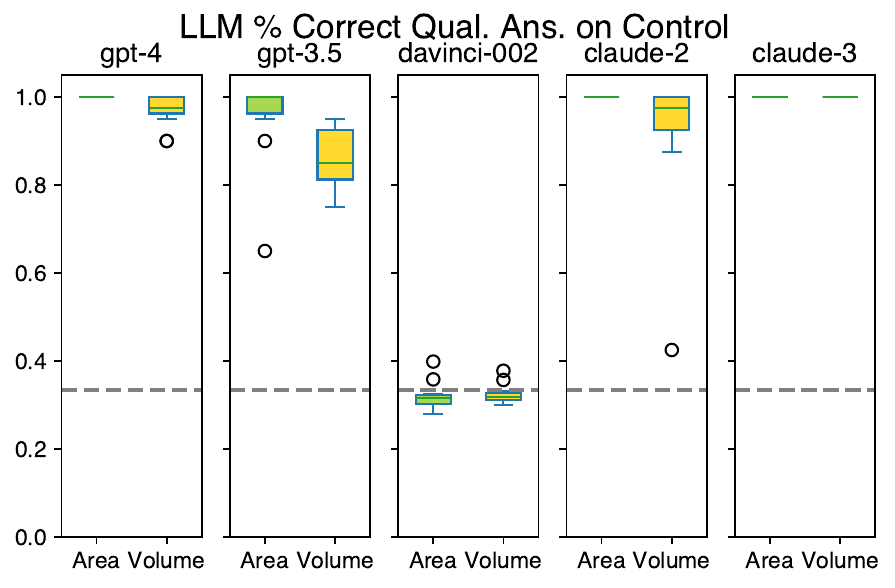}
\caption{Performance of LLMs on the qualification task when the scenarios prompted were ones in which the Nash Product and Utilitarian Sum \textbf{agreed}. Box plots show the average agreement with the correct answer. In the \texttt{Area} condition, models are prompted to choose the proposal which computes the Utilitarian Sum---maximizes the area of the proposals. In the \texttt{Volume} condition, models are prompted to choose the proposal which computes the Nash Product---maximizes the product of the proposals. For prompts see Fig. \ref{fig:claude-qualification}.}
\label{app:llm-qual-control}
\end{figure}

\begin{figure}[tbhp]
\centering
\includegraphics[width=0.5\textwidth]{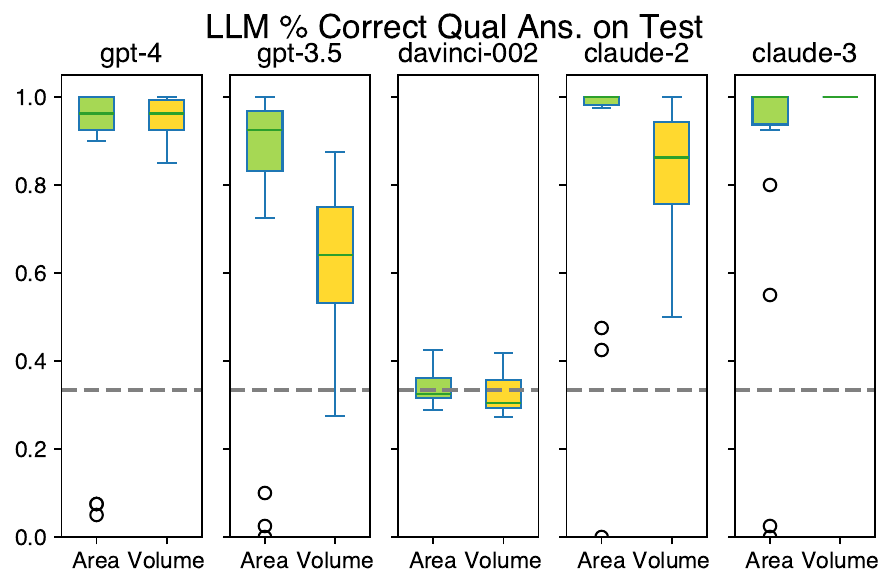}
\caption{Performance of LLMs on the qualification task when the scenarios prompted were ones in which the Nash Product and Utilitarian Sum \textbf{disagreed}. Box plots show the average agreement with the correct answer. In the \texttt{Area} condition, models are prompted to choose the proposal which computes the Utilitarian Sum---maximizes the area of the proposals. In the \texttt{Volume} condition, models are prompted to choose the proposal which computes the Nash Product---maximizes the product of the proposals. For prompts see Fig. \ref{fig:claude-qualification}.}
\label{app:llm-qual-test}
\end{figure}



\begin{figure}[tbhp]
\centering
\includegraphics[width=0.5\textwidth]{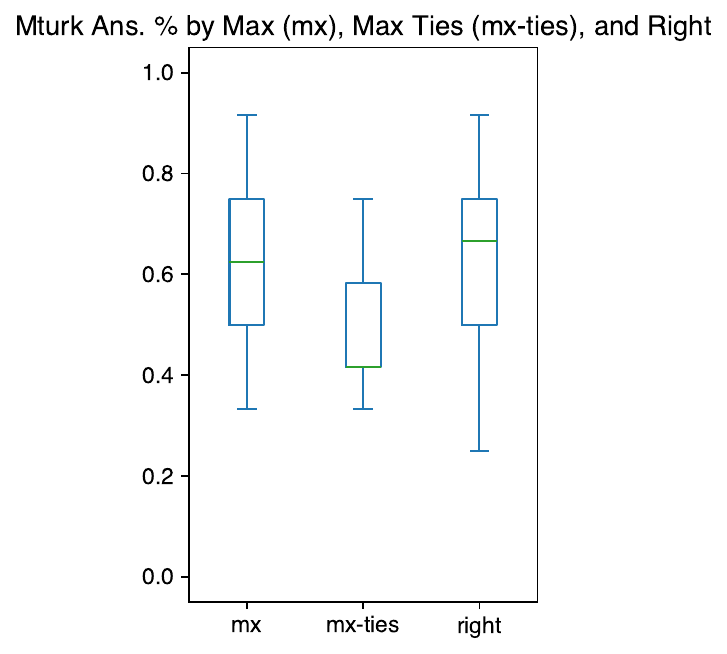}
\caption{The distribution of the support for human participants for the max answer (\texttt{mx}, which measures how often they are in agreement), the max answer under the tie cases (\texttt{mx-ties}), and the correct answer (\texttt{right}).}
\label{app:mturk-max-ties-right}
\end{figure}

\begin{figure}[tbhp]
\centering
\includegraphics[width=0.5\textwidth]{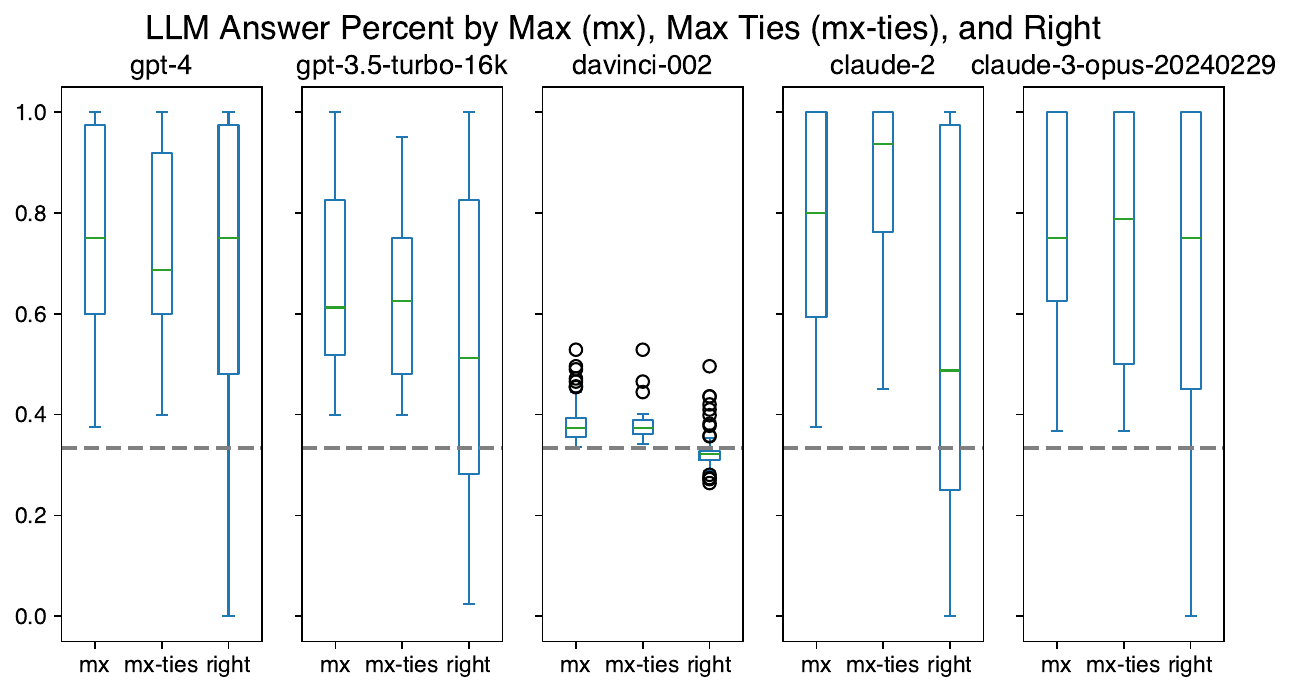}
\caption{The distribution of the support for each model for the max answer (\texttt{mx}, which measures how often each is in agreement with itself), the max answer under the tie cases (\texttt{mx-ties}), and the correct answer (\texttt{right}).}
\label{app:llm-max-tie-right}
\end{figure}

\begin{figure}
\centering
\begin{tabular}{r || r | c | c | c | c}
 &  & \multicolumn{4}{c}{Condition (\# of agreeing responses / total \#)} \\ 
Model &  & \texttt{area} & \texttt{volume} & \texttt{both} & \texttt{none} \\ \hline \hline 
\texttt{gpt-4} & $\Pi$ & 60 / 72*** & 67 / 72*** & 65 / 72*** & 63 / 72*** \\
& $\Sigma$ & 0 / 72*** & 0 / 72*** & 0 / 72*** & 0 / 72*** \\
& $\Sigma$\&$\Pi$ & 18 / 40 & 26 / 40*** & 18 / 40 & 16 / 40 \\ \hline

\texttt{gpt-3.5} & $\Pi$ & 30 / 72 & 48 / 72*** & 43 / 72*** & 53 / 72*** \\
& $\Sigma$ & 18 / 72 & 7 / 72*** & 5 / 72*** & 1 / 72*** \\
& $\Sigma$\&$\Pi$ & 11 / 40 & 23 / 40** & 17 / 40 & 12 / 40 \\ \hline

\texttt{davinci-002} & $\Pi$ & 20 / 72 & 20 / 72 & 20 / 72 & 20 / 72 \\
& $\Sigma$ & 28 / 72 & 28 / 72 & 28 / 72 & 28 / 72 \\
& $\Sigma$\&$\Pi$ & 8 / 40 & 8 / 40 & 8 / 40 & 8 / 40 \\ \hline

\texttt{claude-2} & $\Pi$ & 63 / 72*** & 66 / 72*** & 69 / 72*** & 68 / 72*** \\
& $\Sigma$ & 4 / 72*** & 2 / 72*** & 1 / 72*** & 0 / 72*** \\
& $\Sigma$\&$\Pi$ & 13 / 40 & 20 / 40* & 11 / 40 & 9 / 40 \\ \hline

\texttt{claude-3} & $\Pi$ & 59 / 69*** & 68 / 72*** & 66 / 72*** & 67 / 72*** \\
& $\Sigma$ & 1 / 69*** & 0 / 72*** & 1 / 72*** & 0 / 72*** \\
& $\Sigma$\&$\Pi$ & 19 / 40 & 28 / 40*** & 17 / 40 & 18 / 40 \\ \hline
\end{tabular}

\caption{
Count and number of scenarios with the Nash Product ($\Pi$) or the Utilitarian Sum ($\Sigma$) for \textbf{LLM} disagreement and agreement cases by condition, whether a model saw the \texttt{area} chart, \texttt{volume} chart, \texttt{both}, or \texttt{none}.  (Data for Fig. \ref{fig:llm-control} and \ref{fig:test-conditions-2} and main text Fig. 2B and 4B.) 
In the agreement cases, we had 18 unique scenarios presented with 4 different contexts each answered by each model for 72 responses ($18 \times 4$) total. Similarly, for the agreement cases we had  44 responses ($11 \times 4 $).
In each case, we run a binomial test with a null hypothesis of random guessing (1/3). $\text{***}: p < .001$; $\text{**}: p < .01$; $\text{*}: p < .05$
}
\label{app:results-llm}
\end{figure}

\begin{figure}
\centering
\begin{tabular}{c | c c c}
\toprule
Inequality Aversion & \# disagreements & \# agreements & ratio (disagreements / agreements) \\
\midrule
0.0 & 216 & 120 & 0.56 \\
0.1 & 96 & 240 & 2.50 \\
0.2 & 0 & 336 & 0.00 \\
0.3 & 0 & 336 & 0.00 \\
0.4 & 24 & 312 & 13.00 \\
0.5 & 72 & 264 & 3.67 \\
0.6 & 72 & 264 & 3.67 \\
0.7 & 96 & 240 & 2.50 \\
0.8 & 96 & 240 & 2.50 \\
0.9 & 96 & 240 & 2.50 \\
1.0 & 72 & 120 & 1.67 \\
\bottomrule
\end{tabular}
\caption{
We reanalyzed the set of default scenarios by classifying each in terms of whether the Nash Product (equation 2) and the Inequality Sum (equation 5) \textit{agreed} or \textit{disagreed}. (See Fig. 5 in the main text.)
We varied the inequality aversion parameter of the Inequality Sum from zero to one, recomputing the disagreements and agreements between each parameterization and the Nash Product.
This covers 28 scenarios shown in four different contexts with three unique responses each ($336 = 28 \times 4 \times 3$), excluding the five scenarios from the 32 original which resulted in ties for either mechanism. 
Notice how no disagreements result for inequality aversion values of $.1$ and $.2$.
}
\label{tab:inequality-aversion}
\end{figure}

\begin{figure*}
\centering
\resizebox{
   \textwidth
}{!}{
\begin{tabular}{l | r | c | c | c | c | c | c}
\textbf{SOURCE} & \textbf{GROUP} & \textbf{CREDENCE} & \multicolumn{5}{c}{\textbf{ACTION and OUTCOME}} \\ \toprule \hline

\textbf{Kaleido} & Value, right, or duty & Relevance & Support & Oppose & Either & &  \\ \hline

& Duty to protect  & .99 & 1.7 & 1.0 & 1.3 & &  \\
& Jake's emotional well-being & &  & & & &  \\

& Jake's girlfriend's right& .99 & 1.0 & 2.0 & 1.0 & & \\
&  to not be deceived & & & & & & \\

& Friendship/loyalty & .99 & 1.7 & 1.0 & 1.3 & & \\
& Duty to respect the boundaries & .98 & 1.0 &
2.0 & 1.0 & & \\
& of others' relationships  &  &  &
 & & & \\
& Emotional well-being & .96 & 1.8 & 1.0 & 1.2 & & \\
& Duty to help friends in need & .96 & 1.9 & 1.0 & 1.1 & & \\
& Right to emotional support & .93 & 1.6 & 1.0 & 1.3 & & \\
& Autonomy & .88 & 1.0 & 2.0 & 1.0 & &\\
& Compassion & .86 & 1.0 & 1.0 & 1.1 & &\\
& Right to privacy & .85 & 1.0 & 2.0 & 1.0 & &\\
& Honesty & .84 & 1.0 & 2.0 & 1.0 & &\\ \midrule \hline

\textbf{NLPositionality} & Age Demographic & \% & It's very bad & It's bad & It's okay & It's good & It's very good \\ \hline
& 10-20            & .43 & .23 & .30 & .15 & .23 & .08 \\
& 20-30            & .35 & 0.0 & .43 & .57 & 0.0 & 0.0 \\ \midrule \hline

\textbf{Moral}& Country             & Pop. (k)& [Passengers -> & Law [Illegal & No. Characters & Species [Pets & \\
\textbf{Machines} & & & Pedestrians] & -> Legal] & [Less -> More] & -> Humans] & \\ \hline

& Italy               & 59119 & 1.0 & 1.3 & 1.4 & 1.6 & \\
& Colombia            & 51789 & 1.0 & 1.2 & 1.4 & 1.2 & \\
& United Arab Emirates& 09403 & 1.0 & 1.2 & 1.4 & 1.5 & \\
& Panama              & 04379 & 1.0 & 1.3 & 1.3 & 1.4 & \\
& Slovenia            & 02200 & 1.0 & 1.3 & 1.4 & 1.5 & \\
\end{tabular}
}

\caption{Examples of various naturalistic data sets set up in our framework in which the Utilitarian Sum and the Nash Product disagree. \\
\textbf{Kaleido}: Comparing ten values, rights, and duties generated by the Value Kaleidoscope project \citep{sorensen_value_2023} for the statement "If Owen's friend Jake is having trouble with his girlfriend, and Owen wants to help Jake feel better, is Owen sending flowers to Jake's girlfriend in Jake's name, hoping to patch things up moral?", Kaleido generates values such as "Duty to protect Jake's emotional well-being" which supports the statement and "Jake's girlfriend's right to not be deceived" which opposes the statement. Under these assumptions the Utilitarian Sum opposes the action and the Nash Product supports it. \\
\textbf{NLPositionality}: Comparing the demographic '20-30' and '10-20' for the statement "children doing risky things" on the NLPositionality data set \citep{santy_nlpositionality_2023}. Here the Utilitarian Sum chooses "It's okay" while the Nash Product chooses "It's bad". \\
\textbf{Moral Machines}: Five countries' UN population \citep{united_nations_department_of_economic_and_social_affairs_population_division_world_2022} and preferred categories from the Moral Machines trolley problem experiment \citep{awad_moral_2018}. In this case, the Utilitarian Sum favors saving more characters overall while the Nash Product favors favoring humans over other species.}
\label{app:naturalistic-values}
\end{figure*}


\clearpage


\begin{figure*}
\centering
{
\tiny
\sffamily
\begin{humanbox}{LLM Prompt}
Human: TASK: In this task we assess how to compromise between different views.\\

Tell us which of the options specified is the best compromise for the given situation.\\

Whether one option is a better compromise than another is up to you. It might be that multiple parties have to accept a slightly worse outcome for themselves in order to best balance the desires of the group.\\

In this version of the task, groups prefer higher outcomes. This means that higher outcomes are better.\\

SCENARIO: Your local health department is looking for advice on a project.\\

To figure out what to do, the health department has researched how the proposals will affect different groups. We won't get into the specifics of the proposals, but none is perfect; some groups prefer some proposals over others.\\

TIP: If you find yourself stuck between different proposals, please go with your gut/intuition.\\

Human: In this scenario, there are 3 groups:\\
- group apple-\apple with 33.0 people in it, \\
- group bee-\bee with 33.0 people in it, and \\
- group cow-\cow with 33.0 people in it. \\

There are 3 proposals, each of which will decrease the average number of days a group member must wait for an appointment by:\\
- proposal one: 101.0 days for group apple-\apple, 1.0 days for group bee-\bee, and 1.0 days for group cow-\cow.\\
- proposal two: 1.0 days for group apple-\apple, 51.0 days for group bee-\bee, and 51.0 days for group cow-\cow.\\
- proposal three: 1.0 days for group apple-\apple, 51.0 days for group bee-\bee, and 1.0 days for group cow-\cow.\\

\{\textbf{Chart Description}\}\\

Which proposal is the best compromise in this situation? You must give exactly one answer.\\
- (A) Proposal one\\
- (B) Proposal two\\
- (C) Proposal three\\
\end{humanbox}
\begin{minipage}{.38\textwidth}
\begin{humanbox}{Area Chart Description}
The total area for a proposal is calculated by summing up the product of group size and proposed outcome for each group.

For proposal one, the total area is 101.0 * 33 (for group "apple-\apple") + 1.0 * 33 (for group "bee-\bee") + 1.0 * 33 (for group "cow-\cow") = 3333.0 + 33.0 + 33.0 = 3399.0.

For proposal two, the total area is 1.0 * 33 (for group "apple-\apple") + 51.0 * 33 (for group "bee-\bee") + 51.0 * 33 (for group "cow-\cow") = 33.0 + 1683.0 + 1683.0 = 3399.0.

For proposal three, the total area is 1.0 * 33 (for group "apple-\apple") + 51.0 * 33 (for group "bee-\bee") + 1.0 * 33 (for group "cow-\cow") = 33.0 + 1683.0 + 33.0 = 1749.0.
\end{humanbox}
\end{minipage}
\begin{minipage}{.6\textwidth}
\begin{areabox}
Assistant: Based on the calculations, proposal one results in the highest total area of 3399, but heavily favors group apple. Proposal two results in an equal total area of 3399, but splits it evenly between groups bee and cow. 

Since the goal is to find a compromise between the groups, proposal two balances the outcomes most evenly. Therefore, the best compromise is proposal two.

The answer is B.
\end{areabox}
\end{minipage}
}
\caption{An example of the survey task given to LLMs.  Response depicted above came from\texttt{Claude-2.1} in the area condition. Each had either the area or volume chart description paced in the braces and bolded section in the question (top, in grey) based on the condition.}
\label{app:claude-example-area}
\end{figure*}

\begin{figure*}
\centering
{
\tiny
\begin{minipage}{.48\textwidth}
\begin{humanbox}{Volume Chart Description}
The total volume for a proposal is calculated by multiplying the proposed outcome raised to the (normalized) group size for each group.

For proposal one, the total volume is 101.0 ** 0.3 (for group "apple-\apple") + 1.0 ** 0.3 (for group "bee-\bee") + 1.0 ** 0.3 (for group "cow-\cow") = 4.7 * 1.0 * 1.0 = 4.7.

For proposal two, the total volume is 1.0 ** 0.3 (for group "apple-\apple") + 51.0 ** 0.3 (for group "bee-\bee") + 51.0 ** 0.3 (for group "cow-\cow") = 1.0 * 3.7 * 3.7 = 13.8.

For proposal three, the total volume is 1.0 ** 0.3 (for group "apple-\apple") + 51.0 ** 0.3 (for group "bee-\bee") + 1.0 ** 0.3 (for group "cow-\cow") = 1.0 * 3.7 * 1.0 = 3.7.
\end{humanbox}
\end{minipage}
\begin{minipage}{.5\textwidth}
\begin{volumebox}
Assistant: Based on the calculations, proposal two has the highest total volume of 13.8. This means it provides the best balance of outcomes for the three groups. 

Proposal one favors group apple heavily at the expense of groups bee and cow. Proposal three favors group bee at the expense of groups apple and cow.

Proposal two provides a compromise where each group gets a moderately good outcome. No group gets an extremely good or extremely bad outcome.

Therefore, the best compromise is proposal two. The answer is B.
\end{volumebox}
\end{minipage}
}
\caption{An example survey task given to LLMs.  Response depicted above came from\texttt{Claude-2.1} in the volume condition, continuing the previous figure, \ref{app:claude-example-area}.}
\label{app:claude-example-volume}
\end{figure*}

\clearpage

\begin{figure*}[tbhp]
\centering
{
\tiny
\sffamily
\begin{minipage}{.9\textwidth}
\begin{humanbox}{LLM Qualification Prompt}
Human: This is a qualification task. Your answers will be compared to our ground truth answers. \\

TASK: In this task we assess how well you can judge various aggregated properties of groups. \\

These charts will show you the numeric outcomes on a few proposals for a few groups. \\

Human: In this scenario, there are 3 groups:\\
- group apple-\apple with 33.0 people in it, \\
- group bee-\bee with 33.0 people in it, and \\
- group cow-\cow with 33.0 people in it.\\

There are 3 proposals:\\
- proposal one: with an outcome of 101.0 for group apple-\apple, 1.0 for group bee-\bee, and 1.0 for group cow-\cow.\\
- proposal two: with an outcome of 1.0 for group apple-\apple, 51.0 for group bee-\bee, and 51.0 for group cow-\cow.\\
- proposal three: with an outcome of 1.0 for group apple-\apple, 51.0 for group bee-\bee, and 1.0 for group cow-\cow.

Which of the proposal(s) takes up the largest total \textbf{area}?\\
- (A) Proposal one\\
- (B) Proposal two\\
- (C) Proposal three\\
\end{humanbox}
\end{minipage}
\begin{minipage}{.49\textwidth}
\begin{areabox}
Assistant: Let's calculate the total area for each proposal:

Proposal 1:\\
- Group apple-\apple: 101.0 * 33.0 = 3333.0\\
- Group bee-\bee: 1.0 * 33.0 = 33.0\\
- Group cow-\cow: 1.0 * 33.0 = 33.0\\
- Total area = 3333.0 + 33.0 + 33.0 = 3399.0\\

Proposal 2:  \\
- Group apple-\apple: 1.0 * 33.0 = 33.0\\
- Group bee-\bee: 51.0 * 33.0 = 1683.0\\
- Group cow-\cow: 51.0 * 33.0 = 1683.0\\
- Total area = 33.0 + 1683.0 + 1683.0 = 3399.0\\

Proposal 3:\\
- Group apple-\apple: 1.0 * 33.0 = 33.0 \\
- Group bee-\bee: 51.0 * 33.0 = 1683.0\\
- Group cow-\cow: 1.0 * 33.0 = 33.0\\
- Total area = 33.0 + 1683.0 + 33.0 = 1749.0\\

Proposal 1 has the largest total area of 3399.0.

Therefore, the answer is (A) Proposal one.
\end{areabox}
\end{minipage}
\begin{minipage}{.49\textwidth}
\begin{volumebox}
Assistant: Let's break this down step-by-step:

- There are 3 groups (apple-\apple, bee-\bee, cow-\cow) each with 33 people \\
- To calculate total volume, we take the proposed outcome raised to the power of the normalized group size for each group, and multiply the volumes.

For proposal 1:\\
- apple-\apple volume: 101\textasciicircum0.3333 = 10\\
- bee-\bee volume: 1\textasciicircum0.3333 = 1 \\
- cow-\cow volume: 1\textasciicircum0.3333 = 1\\
- Total volume = 10 * 1 * 1 = 10\\

For proposal 2: \\
- apple-\apple volume: 1\textasciicircum0.3333 = 1\\
- bee-\bee volume: 51\textasciicircum0.3333 = 3.57 \\
- cow-\cow volume: 51\textasciicircum0.3333 = 3.57\\
- Total volume = 1 * 3.57 * 3.57 = 12.7\\

For proposal 3:\\
- apple-\apple volume: 1\textasciicircum0.3333 = 1  \\
- bee-\bee volume: 51\textasciicircum0.3333 = 3.57\\
- cow-\cow volume: 1\textasciicircum0.3333 = 1\\
- Total volume = 1 * 3.57 * 1 = 3.57\\

Proposal 2 has the largest total volume.

Therefore, the answer is B.
\end{volumebox}
\end{minipage}
}
\caption{An example qualification task as asked to \texttt{Claude-2.1} in the area condition (on the bottom left) and the volume condition (on the bottom right), where the bolded word in the question (top, in grey) changes based on the condition. Notice that the area answer is correct, and the math is right. The volume answer is correct, although the math is wrong (e.g., $51.0^{33.0/99.0} = 3.7$).}
\label{fig:claude-qualification}
\end{figure*}



\clearpage
\subsubsection*{Mturk Survey}
\label{app:survey}
\begin{minipage}{\textwidth}
\includepdf[pages=-,scale=0.8, nup=2x3, frame=false, delta=1 1, column=true, pagecommand={}]{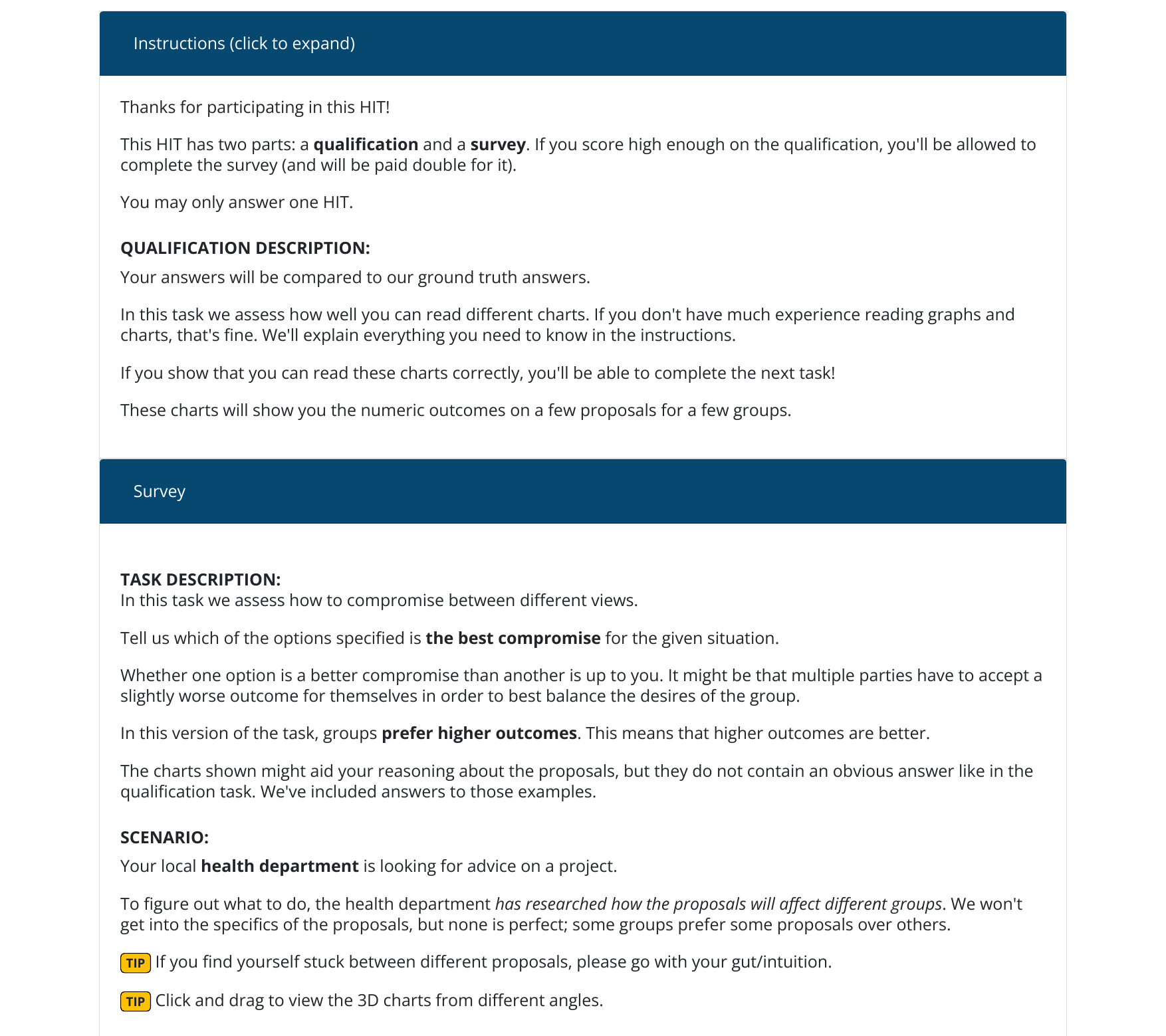}
\end{minipage}
\clearpage

\clearpage

\subsubsection*{Qualification Task}
\label{app:qualification}

\begin{minipage}{\textwidth}
\includepdf[pages=1-10,scale=0.85, nup=2x5, frame=false, delta=1 1, column=true, pagecommand={}]{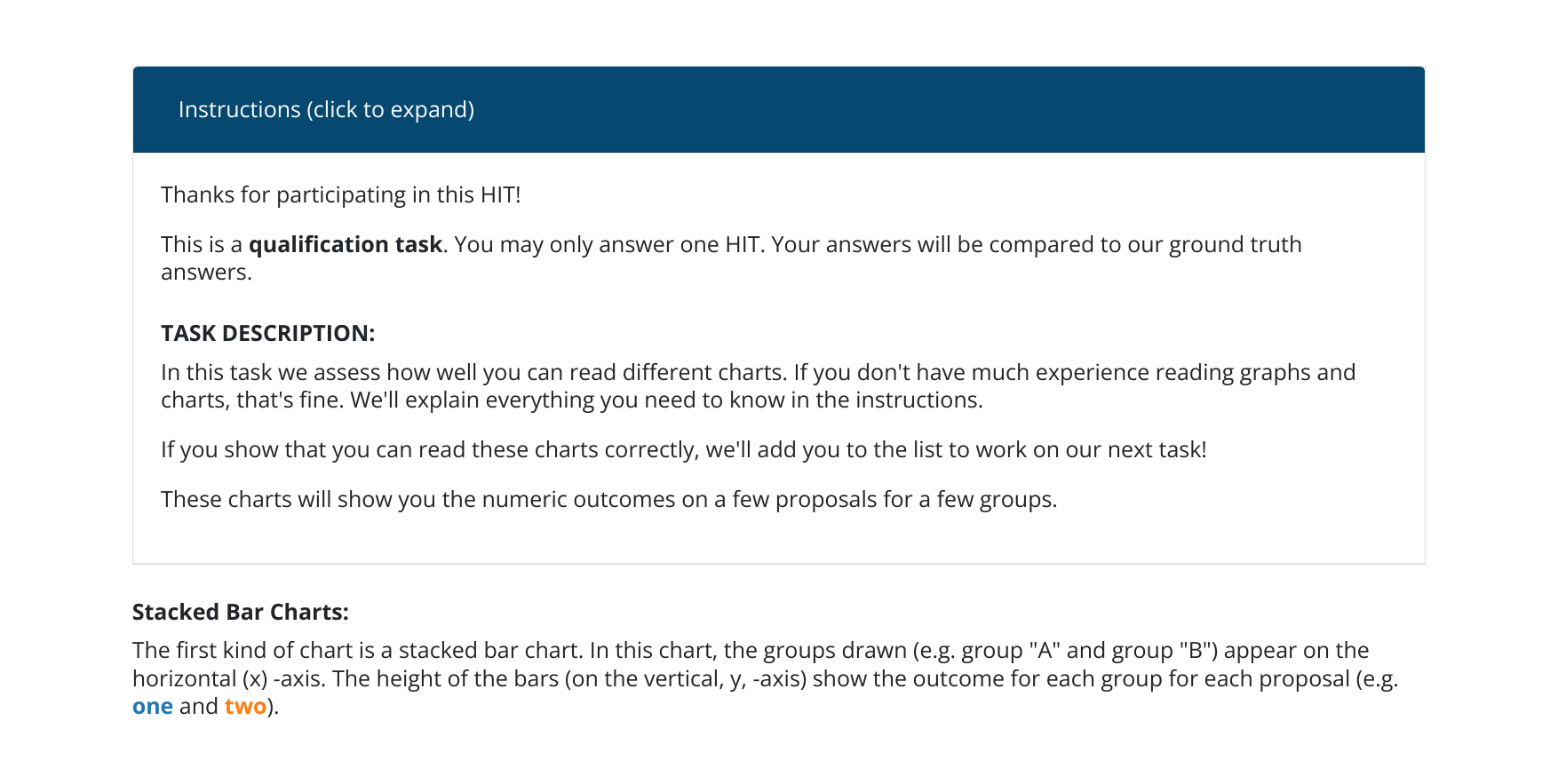}
\end{minipage}
\clearpage
\begin{minipage}{\textwidth}
\includepdf[pages=11-,scale=0.85, nup=2x5, frame=false, delta=1 1, column=true, pagecommand={}]{figures/qualification.pdf}
\end{minipage}
\clearpage
